\documentclass[letterpaper]{article} 
\usepackage{aaai24}  
\usepackage{times}  
\usepackage{helvet}  
\usepackage{courier}  
\usepackage[hyphens]{url}  
\usepackage{graphicx} 
\usepackage{natbib}  
\usepackage{caption} 
\usepackage{algorithm}
\usepackage{algorithmic}
\usepackage[T1]{fontenc} 

\usepackage{amsmath}
\usepackage{amssymb}
\usepackage{mathtools}
\usepackage{amsthm}

\usepackage{multirow}
\usepackage{graphicx}
\usepackage{subfigure}
\usepackage{booktabs}

\usepackage{newfloat}
\usepackage{listings}
\DeclareCaptionStyle{ruled}{labelfont=normalfont,labelsep=colon,strut=off} 
\floatstyle{ruled}
\newfloat{listing}{tb}{lst}{}
\floatname{listing}{Listing}
\title{Wasserstein Differential Privacy}
\author{
   	Chengyi Yang\textsuperscript{\rm 1},
    Jiayin Qi\textsuperscript{\rm 2\thanks{Corresponding Author}},
    Aimin Zhou\textsuperscript{\rm 1}
}
\affiliations{
    \textsuperscript{\rm 1}Shanghai Institute of AI for Education, School of Computer Science and Technology, and Key Laboratory of MEA (Ministry of Education), 
    East China Normal University\\
    \textsuperscript{\rm 2}Cyberspace Institute of Advanced Technology, Guangzhou University\\
    52265901027@stu.ecnu.edu.cn, qijiayin@139.com, amzhou@cs.ecnu.edu.cn
}

\usepackage{bibentry}

\begin{document}

\maketitle

\begin{abstract}
Differential privacy (DP) has achieved remarkable results in the field of privacy-preserving machine learning. However, existing DP frameworks do not satisfy all the conditions for becoming metrics, which prevents them from deriving better basic private properties and leads to exaggerated values on privacy budgets. We propose Wasserstein differential privacy (WDP), an alternative DP framework to measure the risk of privacy leakage, which satisfies the properties of symmetry and triangle inequality. We show and prove that WDP has 13 excellent properties, which can be theoretical supports for the better performance of WDP than other DP frameworks. 
In addition, we derive a general privacy accounting method called Wasserstein accountant, which enables WDP to be applied in stochastic gradient descent (SGD) scenarios containing subsampling. Experiments on basic mechanisms, compositions and deep learning show that the privacy budgets obtained by Wasserstein accountant are relatively stable and less influenced by order. Moreover, the overestimation on privacy budgets can be effectively alleviated. The code is available at \url{https://github.com/Hifipsysta/WDP}. 
\end{abstract}

\section{Introduction}\label{section:introduction}

Differential privacy \cite{DBLP:conf/tcc/DworkMNS06} is a mathematically rigorous definition of privacy, providing quantifiable descriptions of the risk on leaking sensitive information. 
In the early stage, researches on differential privacy mainly focused on the issue of statistical queries (SQ) \cite{DBLP:conf/sigmod/McSherry09, DBLP:journals/siamcomp/KasiviswanathanLNRS11}. 
With the risk of privacy leakage being warned in machine learning \cite{DBLP:conf/ijcai/0009SW15,DBLP:conf/sp/ShokriSSS17,DBLP:conf/nips/ZhuLH19}, differential privacy has been gradually applied for privacy protection in deep learning \cite{DBLP:conf/ccs/ShokriS15, DBLP:conf/ccs/AbadiCGMMT016, DBLP:conf/ijcai/PhanVLJDWT19, DBLP:conf/aaai/ChengWZCW022}. 

However, these techniques are always constructed on the postulation of standard DP~\citep{DBLP:conf/tcc/DworkMNS06}, which only provides the worst-case scenario, and tends to overestimate privacy budgets under the measure of maximum divergence \citep{DBLP:conf/icml/TriastcynF20}. Although the most commonly applied approximate differential privacy ($\varepsilon,\delta$-DP)~\cite{DBLP:conf/eurocrypt/DworkKMMN06} ignores extreme situations with small probabilities by introducing a relaxation term $\delta$ called failure probability, it is believed that $(\varepsilon,\delta)$-DP cannot strictly handle composition problems \cite{DBLP:conf/csfw/Mironov17, 10.1111/rssb.12454}. 
To address the above issues, further researches have been considering the specific data distribution, which can be divided into two main directions: \textit{the distribution of privacy loss} and \textit{the distribution of unique difference}. For example, concentrated differential privacy (CDP)~\cite{DBLP:journals/corr/DworkR16}, zero-concentrated differential privacy (zCDP)~\cite{DBLP:conf/tcc/BunS16}, and truncated concentrated differential privacy (tCDP)~\cite{DBLP:conf/stoc/BunDRS18} all assume that the mean of privacy loss follows subgaussian distribution. While Bayesian differential privacy (BDP)~\cite{DBLP:conf/icml/TriastcynF20}  considers the distribution of the only different data entry $x'$. Nevertheless, they are all defined by the upper bound of divergence, which implies that their privacy budgets are overly pessimistic~\cite{DBLP:conf/icml/TriastcynF20}. 

In this paper, we introduce a variant of differential privacy from another perspective. We define the privacy budget through the upper bound of the Wasserstein distance between adjacent distributions, which is called \textit{Wasserstein differential privacy} (WDP). 
From a semantic perspective, WDP also follows the concept of indistinguishability~\cite{DBLP:conf/tcc/DworkMNS06} in differential privacy. Specifically, for all possible adjacent databases $D$ and $D'$, WDP reflects the maximum variation of optimal transport (OT) cost between the distributions queried by an adversary before and after any data entry change in the database. 

Intuitively speaking, the advantages of WDP can be divided into at least two aspects. (1) WDP focuses on individuals within the distribution, rather than focusing on the entire distribution like divergence, which is consistent with the original intention of differential privacy to protect individual private information from leakage. (2) More importantly, WDP satisfies all the conditions to become a metric, including non-negativity, symmetry and triangle inequality (see Proposition 1-3), which is not fully possessed by privacy loss under the definition of divergence, as divergence itself does not satisfy symmetry and triangle inequality (see Proposition 11 in the appendix of \citet{DBLP:conf/csfw/Mironov17}).

The combination of DP and OT has been taken into consideration in several existing works. Their contributions are essentially to provide privacy guarantees for computing Wasserstein distance between data domains \cite{DBLP:conf/ijcai/LeTienHS19}, distributions \cite{DBLP:conf/icml/RakotomamonjyR21} or graph embeddings \cite{DBLP:conf/ijcai/JinC22}. However, our work is to compute privacy budgets through Wasserstein distance, and the contributions are summarized as follows: 

Firstly, we propose an alternative DP framework called Wasserstein differential privacy (WDP), which satisfies three basic properties of a metric (non-negativity, symmetry and triangle inequality), 
and is easy to convert with other DP frameworks (see Proposition 9-11). 

Secondly, we show that WDP has 13 excellent properties. More notably, basic sequential composition, group privacy among them and advanced composition are all derived from triangle inequality, which shows the advantages of WDP as a metric DP.

Thirdly, we derive advanced composition, privacy loss and absolute moment under WDP, and finally develop Wasserstein accountant to track and account privacy budgets in subsampling algorithms such as SGD in deep learning.

Fourthly, we conduct experiments to evaluate WDP on basic mechanisms, compositions and deep learning. Results show that applying WDP as privacy framework can effectively avoid overstating the privacy budgets. 

\section{Related Work}
Pure differential privacy ($\varepsilon$-DP)~\cite{DBLP:conf/tcc/DworkMNS06} provides strict guarantees for all measured events through maximum divergence. To address the long tailed distribution generated by privacy mechanism, $(\varepsilon,\delta)$-DP~\cite{DBLP:conf/eurocrypt/DworkKMMN06} ignores extremely low probability events through a relaxation term $\delta$. However, $(\varepsilon,\delta)$-DP is considered to an overly relaxed definition~\citep{DBLP:conf/stoc/BunDRS18} and cannot effectively handle composition problems, such as leading to parameter explosion~\cite{DBLP:conf/csfw/Mironov17} or failing to capture correct hypothesis testing~\cite{10.1111/rssb.12454}. In view of this, CDP~\cite{DBLP:journals/corr/DworkR16} applies a subgaussian assumption to the mean of privacy loss. zCDP~\cite{DBLP:conf/tcc/BunS16} capture privacy loss is a subgaussian random variable through Rényi divergence. Rényi differential privacy (RDP)~\cite{DBLP:conf/csfw/Mironov17} proposes a more general definition of DP based on Rényi divergence. tCDP~\cite{DBLP:conf/stoc/BunDRS18} further relaxes zCDP. BDP~\cite{DBLP:conf/icml/TriastcynF20} considers the distribution of unique different entries. Subspace differential privacy~\cite{DBLP:conf/aaai/0001GY22} and integer subspace differential privacy~\cite{DBLP:conf/aaai/Dharangutte0GY23} consider privacy computing scenarios with external constraints. 
However, these concepts are all based on divergence, so that their privacy loss does not have the property of metrics.
Although $f$-DP and its special case Gaussian differential privacy (GDP)~\cite{10.1111/rssb.12454} innovatively define privacy based on the trade-off function between two types of errors in hypothesis testing, they are difficult to associate with other DP frameworks.

\section{Wasserstein Differential Privacy} \label{section:Wasserstein_differential_privacy}

In this section, we introduce the concept of Wasserstein distance and define our Wasserstein differential privacy.

\noindent\textbf{Definition 1} (Wasserstein distance \cite{Ludger2009Optimal}).  For two probability distributions $P$ and $Q$ defined over $\mathcal{R}$, their $\mu$-Wasserstein distance is
\begin{equation}
	\begin{aligned}
		W_\mu\left(P,Q\right)=\left(\inf_{\gamma\in\Gamma\left(P,Q\right)}\int_{\mathcal{X}\times \mathcal{Y}}{{\rho\left(x,y\right)}^\mu d\gamma\left(x,y\right)}\right)^\frac{1}{\mu}
	\end{aligned}.
	\label{equation:wasserstein_distance:definition}
\end{equation}
Where $\rho(x,y)=\Vert x-y \Vert$ is the norm defined in probability space $\Omega = \mathcal{X}\times \mathcal{Y}$. 
$\Gamma\left(P,Q\right)$ is the set for all the possible joint distributions, and  $\gamma(x,y)>0$ satisfying $\int\gamma\left(x,y\right)dy=P(x)$ and $\int\gamma\left(x,y\right)dx=Q(y)$.

In practical sense, $\rho\left(x,y\right)$ can be regarded as the cost for one unit of mass transported from $x$ to $y$. $\gamma\left(x,y\right)$ can be seen as a transport plan representing the share to be moved from $P$ to $Q$, which measures how much mass must be transported in order to complete the transportation.


In particular, when $\mu$ is equal to 1, we can obtain the 1-Wasserstein distance applied in Wasserstein generative adversarial network (WGAN) \cite{DBLP:conf/icml/ArjovskyCB17, DBLP:conf/nips/GulrajaniAADC17}. The successful application of 1-Wasserstein distance in WGAN should be attributed to Kantorovich-Rubinstein duality, which effectively reduces the computational complexity of Wasserstein distance. 

\noindent\textbf{Definition 2} (Kantorovich-Rubinstein distance \cite{1958On}). According to the property of Kantorovich-Rubinstein duality, 1-Wasserstein distance can be equivalently expressed as Kantorovich-Rubinstein distance
\begin{equation}
	K\left(P,Q\right)=\sup_{\Vert \varphi \Vert_{L} \le 1} \mathbb{E}_{x\sim P} [\varphi(x)]-\mathbb{E}_{y\sim Q} [\varphi(y)].
\end{equation}
Where $\varphi:\mathcal{X}\rightarrow \mathcal{R}$ is the so-called Kantorovich potential, giving the optimal transport map by a close-form formula. Where $\Vert \varphi \Vert_{L}$ is the Lipschitz bound of Kantorovich potential, $\Vert \varphi\Vert_{L}\le 1$ indicates that $\varphi$ satisfies the 1-Lipschitz condition with
\begin{equation}
	\Vert \varphi \Vert_{L} = \sup_{x\not= y} \frac{ \rho\left(\varphi(x), \varphi(y)\right)}{\rho\left( x, y\right)}.
\end{equation}

\noindent\textbf{Definition 3} ($\left(\mu,\varepsilon\right)$-WDP). A randomized algorithm $\mathcal{M}$ is said to satisfy $\left(\mu,\varepsilon\right)$-Wasserstein differential privacy if for any adjacent datasets $D,D'\in \mathcal{D}$ and all measurable subsets $S\subseteq\mathcal{R}$ the following inequality holds
\begin{equation}
	\begin{aligned}
		&W_\mu\left(Pr[{\mathcal{M}\left({D}\right)\in S}], Pr[{\mathcal{M}\left({D}^\prime\right)}\in S]\right)
		=\\
		&
		\left(\inf_{\gamma\in\Gamma\left(
			{Pr_\mathcal{M}\left({D}\right)},{Pr_\mathcal{M}\left({D}^\prime\right)}
			\right)}\int_{\mathcal{X}\times \mathcal{Y}}{{\rho\left(x,y\right)}^\mu d\gamma\left(x,y\right)}\right)^\frac{1}{\mu}
		\le
		\varepsilon. 
	\end{aligned}\label{WDP_definition}
\end{equation}
Where $\mathcal{M}(D)$ and $\mathcal{M}(D')$ represent two outputs when algorithm $\mathcal{M}$ respectively performs on dataset $D$ and $D'$. $Pr[{\mathcal{M}\left({D}\right)\in S}]$ and $Pr[{\mathcal{M}\left({D'}\right)\in S}]$ are the probability distributions, also denoted as $Pr_\mathcal{M}(D)$ and $Pr_\mathcal{M}(D')$ in this paper. 
The value of $W_\mu\left({Pr_\mathcal{M}\left({D}\right)},{Pr_\mathcal{M}\left({D}^\prime\right)}\right)$ is the privacy loss under $\left(\mu,\varepsilon\right)$-WDP and its upper bound $\varepsilon$ is called privacy budget. 

\noindent\textbf{Symbolic representations.} WDP can also be represented as
$W_\mu(\mathcal{M}(D), \mathcal{M}(D'))\leq \varepsilon$. To emphasize the inputs are two probability distributions, we denote WDP as $W_\mu(Pr_\mathcal{M}(D), Pr_\mathcal{M}(D'))\leq \varepsilon$. To avoid confusion, we also represent RDP as $D_\alpha(Pr_\mathcal{M}(D)\Vert Pr_\mathcal{M}(D'))\leq \varepsilon$, although the representation $D_\alpha(\mathcal{M}(D)\Vert \mathcal{M}(D'))\leq\varepsilon$ implies that the results depend on the randomized algorithm and the queried data. 
They are both reasonable because $\mathcal{M}(D)$ can be seen as a random variable that satisfies $\mathcal{M}(D)\sim Pr_{\mathcal{M}}(D)$. 

For the convenience on computation, we define Kantorovich Differential Privacy (KDP) as an alternative way to obtain privacy loss or privacy budget under $(1,\varepsilon)$-WDP.

\noindent \textbf{Definition 4} (Kantorovich Differential Privacy). If a randomized algorithm $\mathcal{M}$ satisfies$\left(1,\varepsilon\right)$-WDP, which can also be written as the form of Kantorovich-Rubinstein duality 
\begin{align}
	&K\left({Pr_\mathcal{M}\left({D}\right)},{Pr_\mathcal{M}\left({D}^\prime\right)}\right) \nonumber=
	\\&
	\sup_{\Vert \varphi\Vert_{L}\le 1}\mathbb{E}_{x\sim {Pr_\mathcal{M}\left({D}\right)}}[\varphi(x)]
	-\mathbb{E}_{x\sim {Pr_\mathcal{M}\left({D'}\right)}}[\varphi(x)]\le\varepsilon.
	\label{equation:Kantorovich_differential_privacy}
\end{align}
$\varepsilon$-KDP is equivalent to $\left(1,\varepsilon\right)$-WDP, and can be computed more efficiently through duality formula based on Kantorovich-Rubinstein distance.

\section{Properties of WDP} \label{section:properties_of_WDP}

\noindent\textbf{Proposition 1} (Symmetry). Let $\mathcal{M}$ be a $\left(\mu,\varepsilon\right)$-WDP algorithm, for any $\mu \ge 1$ and $\varepsilon \ge 0$ the following equation holds
\begin{equation}
	\begin{aligned}
		W_\mu  (Pr_\mathcal{M}\left({D}\right), & Pr_\mathcal{M}\left({D}^\prime\right))
		\\
		=&W_\mu\left(Pr_\mathcal{M}\left(D^\prime\right),Pr_\mathcal{M}\left(D\right)\right)\le\varepsilon.
	\end{aligned}
\end{equation}
The symmetric property of $\left(\mu,\varepsilon\right)$-WDP is implied in its definition. Specifically, the joint distribution $\Gamma(\cdot)$ satisfies $\Gamma(Pr_\mathcal{M}(D^\prime),Pr_\mathcal{M}(D))=\Gamma(Pr_\mathcal{M}(D),Pr_\mathcal{M}(D^\prime))$. 
In addition, Kantorovich differential privacy also satisfies this property and the proof is available in the appendix.


\noindent \textbf{Proposition 2}  (Triangle Inequality) Let $D_1,D_2,D_3\in \mathcal{D}$ be three arbitrary datasets. Suppose there are fewer different data entries between $D_1$ and $D_2$ compared with $D_1$ and $D_3$, and the differences between $D_1$ and $D_2$ are included in the differences between $D_1$ and $D_3$. For any randomized algorithm $\mathcal{M}$ satisfies $(\mu, \varepsilon)$-WDP with $\mu \ge 1$, we have
\begin{equation}
	\begin{aligned}
		W_{\mu}(Pr_\mathcal{M}(D_1),&Pr_\mathcal{M}(D_3))
		\\
		\le &W_{\mu}(Pr_\mathcal{M}(D_1),Pr_\mathcal{M}(D_2))\\
		&+W_{\mu}(Pr_\mathcal{M}(D_2),Pr_\mathcal{M}(D_3)).
	\end{aligned}\label{equation:triangle_inequality}
\end{equation}

The proof is available in the appendix, and Minkowski's inequality is applied in the deduction process. Proposition 2 can also be understood as the cost that converting from $Pr_\mathcal{M}(D_1)$ to $Pr_\mathcal{M}(D_2)$ and then to $Pr_\mathcal{M}(D_3)$ is not lower than the cost that converting from $Pr_\mathcal{M}(D_1)$ to $Pr_\mathcal{M}(D_3)$ directly. Triangle inequality is indispensable in proving several properties, such as basic sequential composition (see Proposition 6), group privacy (see Proposition 13) and advanced composition (see Theorem 1).

\noindent\textbf{Proposition 3} (Non-Negativity). For $\mu\ge 1$ and any randomized algorithm $\mathcal{M}$, we have $W_{\mu}(Pr_\mathcal{M}(D), Pr_\mathcal{M}(D'))\ge 0$. 

\noindent\textit{Proof.}  See proof of Proposition 3 in the appendix.

\noindent \textbf{Proposition 4} (Monotonicity). For $1 \le \mu_1\le \mu_2$, we have $W_{\mu_1}(Pr_\mathcal{M}(D),Pr_\mathcal{M}(D'))\le W_{\mu_2}(Pr_\mathcal{M}(D),Pr_\mathcal{M}(D'))$, or we can equivalently described this proposition as $(\mu_2,\varepsilon)$-WDP implies $(\mu_{1},\varepsilon)$-WDP.

The proof is available in the appendix, and the derivation is completed with the help of Lyapunov's inequality. 


\noindent\textbf{Proposition 5} (Parallel Composition). Suppose a dataset $D$ is divided into $n$ parts disjointly which are denoted as $D_{i}, i=1,2,\cdots,n$. Each randomized algorithm $\mathcal{M}_i$  performed on different seperated datasets $D_i$ respectively. If $\mathcal{M}_i:\mathcal{D}\rightarrow\mathcal{R}_i$ satisfies $\left(\mu,\varepsilon_i\right)$-WDP for $i=1,2,\cdots,n$, then the set of  randomized algorithms  $\mathcal{M}=\{\mathcal{M}_1,\mathcal{M}_2,\cdots,\mathcal{M}_n\}$ satisfies ($\mu$, $\max\{\varepsilon_1,\varepsilon_2,\cdots,\varepsilon_n\}$)-WDP.

\noindent\textit{Proof.} See proof of Proposition 5 in the appendix.

\noindent \textbf{Proposition 6} (Sequential Composition). Consider a series of randomized algorithms $\mathcal{M}=\{\mathcal{M}_1,\cdots, \mathcal{M}_i, \cdots, \mathcal{M}_n\}$  performed on a dataset sequentially. If any $\mathcal{M}_i:\mathcal{D}\rightarrow\mathcal{R}_i$ satisfies $\left(\mu,\varepsilon_i\right)$-WDP, then $\mathcal{M}$ satisfies $(\mu, \sum_{i=1}^{n}\varepsilon_i)$-WDP.

\noindent\textit{Proof.} See proof of Proposition 6 in the appendix. 

\noindent\textbf{Proposition 7} (Laplace Mechanism). If an algorithm $f:\mathcal{D}\rightarrow\mathcal{R}$ has sensitivity $\Delta_p f$ and the order $\mu\geq1$, then the Laplace mechanism $\mathcal{M}_L=f\left(x\right)+Lap\left(0,\lambda\right)$  preserves $\left(\mu, \frac{1}{2} \Delta_p f \left(\sqrt{2\left[ 1/\lambda + \exp(-1/\lambda)-1 \right]} \right)^{\frac{1}{\mu}} \right)$-WDP. 

\noindent\textit{Proof.} See proof of Proposition 7 in the appendix. 

\noindent\textbf{Proposition 8} (Gaussian Mechanism). If an algorithm $f:\mathcal{D}\rightarrow\mathcal{R}$ has sensitivity $\Delta_p f$ and the order $\mu\geq 1$,  then the Gaussian mechanism $\mathcal{M}_G=f\left(x\right)+\mathcal{N}\left(0,\sigma^2\right)$ preserves
$\left(\mu, \frac{1}{2} \left({\Delta_p f}/{\sigma} \right)^{\frac{1}{\mu}} \right)$-WDP.

The proof of Gaussian mechanism is available in the appendix. The relation between parameters and privacy budgets in Laplace mechanism and Gaussian mechanism are summarized in Table \ref{parameters_and_privacy_budgets}.


\begin{table*}[htb]
	\begin{center}   
		\begin{tabular}{ccc}   
			\toprule   \textbf{Differential Privacy Framework} & \textbf{Laplace Mechanism} & \textbf{Gaussian Mechanism} \\   
			\midrule   
			\multirow{2}{*}{DP}  & \multirow{2}{*}{1/$\lambda$} &   \multirow{2}{*}{$\infty$} \\ 
			&&\\
			\midrule   
			\multirow{3}{*}{RDP for order $\alpha$}& $\alpha>1$: $\frac{1}{\alpha-1}\log\left\{\frac{\alpha}{2\alpha-1} \exp\left( \frac{\alpha-1}{\lambda} 
			\right) +
			\frac{\alpha-1}{2\alpha-1} \exp \left(-\frac{\alpha}{\lambda}\right)
			\right\}$   & \multirow{3}{*}{$\alpha/(2\sigma^2)$}  \\  
			&&\\
			& $\alpha=1$: $1/\lambda+\exp\left(-1/\lambda\right)-1$  & \\  
			\midrule     
			\multirow{2}{*}{WDP for order $\mu$}&\multirow{2}{*}{$\frac{1}{2}\Delta_p f\left(\sqrt{2\left[ 1/\lambda + \exp(-1/\lambda)-1 \right]} \right)^{\frac{1}{\mu}}$}&\multirow{2}{*}{$\frac{1}{2} \left({\Delta_p f}/{\sigma} \right)^{\frac{1}{\mu}} $} \\
			&&\\
			\bottomrule 
		\end{tabular}   
	\end{center}   
	\caption{Privacy budgets of DP, RDP and WDP for Basic Mechanisms. The Laplace mechanism and Gaussian mechanism of DP and RDP with sensitivity 1 are obtained from Table 2 in \citet{DBLP:conf/csfw/Mironov17}. When it comes to WDP, the sensitivity $\Delta_p f$ can be an arbitrary positive constant.}  
	\label{parameters_and_privacy_budgets} 
\end{table*}

\noindent\textbf{Proposition 9} (From DP to WDP) If $\mathcal{M}$ preserves $\varepsilon$-DP with sensitivity $\Delta f$
, it also satisfies $\left(\mu, \frac{1}{2} \Delta_p f \left({2 \varepsilon \cdot (e^\varepsilon-1) }\right)^{\frac{1}{2 \mu}} \right)$-WDP.

\noindent\textit{Proof.} See proof of Proposition 9 in the appendix. 


\noindent \textbf{Proposition 10} (From RDP to WDP) If $\mathcal{M}$ preserves $(\alpha,\varepsilon)$-RDP with sensitivity $\Delta_p f$, it also satisfies $\left(\mu,\frac{1}{2} \Delta_p f  \left(2\varepsilon\right)^{\frac{1}{2\mu}} \right)$-WDP.

\noindent\textit{Proof.} See proof of Proposition 10 in the appendix. 





\noindent \textbf{Proposition 11} (From WDP to RDP and DP) Suppose $\mu \geq 1$ and $\log(p_\mathcal{M}(\cdot))$ is an $L$-Lipschitz function. 
If $\mathcal{M}$ preserves $(\mu,\varepsilon)$-WDP with sensitivity $\Delta_p f$, it also satisfies $\left(\alpha, \frac{\alpha}{\alpha-1}  L  \cdot \varepsilon^{\mu/(\mu+1)} \right)$-RDP.
Specifically, when $\alpha \rightarrow \infty$, $\mathcal{M}$ satisfies $\left(L  \cdot \varepsilon^{\mu/(\mu+1)}\right)$-DP. 

The proof is available in the appendix. Where $p_\mathcal{M}(\cdot)$ is the probability density function of distribution $Pr_{\mathcal{M}}(\cdot)$.

\noindent\textbf{Proposition 12} (Post-Processing). Let $\mathcal{M}:\mathcal{D}\rightarrow \mathcal{R}$ be a $(\mu,\varepsilon)$-Wasserstein differentially private algorithm. Let $\mathcal{G}: \mathcal{R} \rightarrow \mathcal{R}'$ be an arbitrary randomized mapping. For any order $\mu\in[1,\infty)$ and all measurable subsets $S\subseteq\mathcal{R}$, $\mathcal{G}(\mathcal{M})(\cdot)$ is also $(\mu,\varepsilon)$-Wasserstein differentially private, namely
\begin{align}
	W_\mu \left(Pr[\mathcal{G}(\mathcal{M}(D))\in S], Pr[\mathcal{G}(\mathcal{M}(D'))\in S]\right)\le \varepsilon.
\end{align}

\noindent\textit{proof.} See proof of Proposition 12 in the appendix.


\noindent\textbf{Proposition 13} (Group Privacy). Let $\mathcal{M} : \mathcal{D} \mapsto \mathcal{R}$ be a $(\mu, \varepsilon)$-Wasserstein differentially private algorithm. Then for any pairs of datasets $D, D' \in \mathcal{D}$ differing in $k$ data entries $x_1,\cdots,x_k$ for any $i = 1, \cdots, k, \mathcal{M}$ is
$(\mu, k \varepsilon)$-Wasserstein differentially private.

\noindent\textit{Proof.} See proof of Proposition 13 in the appendix.

\section{Implementation in Deep Learning}

\subsection{Advanced Composition}

To derive advanced composition under WDP, we first define generalized $\left(\mu,\varepsilon\right)$-WDP.

\noindent\textbf{Definition 5} (Generalized $\left(\mu,\varepsilon\right)$-WDP) 
A randomized mechanism $\mathcal{M}$ is generalized  $\left(\mu,\varepsilon\right)$-Wasserstein differentially private if for any two adjacent datasets $D,D^\prime\in\mathcal{D}$ holds that
\begin{equation}
	Pr[W_\mu(Pr_\mathcal{M}(D),Pr_\mathcal{M}(D'))\ge \varepsilon] \le \delta. \label{equation:definition_for_generalized_RDP}
\end{equation}
According to the above definition, we find that
$(\mu,\varepsilon)$-WDP can be regarded as a special case of generalized $(\mu,\varepsilon)$-WDP when $\delta$ tends to zero.

Definition 5 is helpful for designing Wasserstein accountant applied in private deep learning, and we will deduce several necessary theorems based on this notion in the following. 

\noindent\textbf{Theorem 1} (Advanced Composition) Suppose a randomized algorithm $\mathcal{M}$ consists of a sequence of $(\mu,\varepsilon)$-WDP algorithms $\mathcal{M}_1, \mathcal{M}_2, \cdots, \mathcal{M}_T$, which perform on dataset $D$ adaptively and satisfy $\mathcal{M}_t: \mathcal{D}\rightarrow \mathcal{R}_t$, $t\in\{1,2,\cdots,T\}$. $\mathcal{M}$ is generalized $(\mu,\varepsilon)$-Wasserstein differentially private with $\varepsilon> 0$ and $\mu\geq1$ if for any two adjacent datasets $D,D^\prime\in\mathcal{D}$ hold that
\begin{equation}\label{equation:Theorem_1}
	{ \exp
		\left[ 
		\beta 
		\sum_{t=1}^T 
		\mathbb{E}(
		W_\mu(Pr_{\mathcal{M}_t}(D),Pr_{\mathcal{M}_t}(D^\prime)))-
		\beta 
		\varepsilon
		\right]
	}    \leq \delta .  
\end{equation}
Where $\beta$ is a customization parameter that satisfies $\beta>0$.

\noindent\textit{Proof.} See proof of Theorem 1 in the appendix. 

\subsection{Privacy Loss and Absolute Moment}

\noindent \textbf{Theorem 2} Suppose an algorithm $\mathcal{M}$ consists of a sequence of private algorithms $\mathcal{M}_1, \mathcal{M}_2, \cdots,\mathcal{M}_T$ protected by Gaussian mechanism and satisfying $\mathcal{M}_t: \mathcal{D} \rightarrow \mathcal{R}$, $t=\left\{1,2,\cdots,T\right\}$. 
If the subsampling probability, scale parameter and $l_2$-sensitivity of algorithm $\mathcal{M}_t$ are represented by $q\in [0,1]$, $\sigma>0$ and $d_t\geq 0$, then the privacy loss under WDP at epoch $t$ is
\begin{align}
	&W_{\mu} \left(Pr_{\mathcal{M}_t}(D), Pr_{\mathcal{M}_t}(D')\right)=
	\inf_{d_t}  \left[ \sum_{i=1}^n 
	\mathbb{E}\left(
	\vert Z_{ti} \vert^\mu
	\right) \right]^\frac{1}{\mu} , \nonumber
	\\ 
	&Z_t \sim\mathcal{N}\left(q d_t,(2-2q+2q^2)\sigma^2\right) .
\end{align}
Where $Pr_{\mathcal{M}_t}(D)$ is the outcome distribution when performing $\mathcal{M}_t$ on $D$ at epoch $t$. $d_t=\Vert g_t - g_t'\Vert_2$ represents the $l_2$ norm between pairs of adjacent  gradients $g_t$ and $g_t'$.
In addition, $Z_t$ is a vector follows Gaussian distribution, and $Z_{ti}$ represents the $i$-th component of $Z_t$.

\noindent\textit{Proof.} See proof of Theorem 2 in the appendix.

Note that $\mathbb{E}\left(\vert
Z_{ti}
\vert^\mu\right)$ is the $\mu$-order raw absolute moment of the Gaussian distribution $\mathcal{N}\left(q d_t,(2-2q+2q^2)\sigma^2\right)$. We know that the raw moment of a Gaussian distribution can be obtained by taking the $\mu$-th order derivatives of the moment generating function with respect to $z$. 
Nevertheless, we do not adopt such an indirect approach. We successfully derive a direct formula, as shown in Lemma 1.

\noindent \textbf{Lemma 1} (Raw Absolute Moment) Assume that $Z_t \sim \mathcal{N}(q  d_t,(2-2q+2q^2) \sigma^2)$, we can obtain the raw absolute moment of $Z$ as follow
\begin{align}
	\mathbb{E} \left(\vert Z_t \vert^\mu \right) = 
	\left( 2 Var\right)^{\frac{\mu}{2}}\frac{
		GF\left({\frac{\mu+1}{2}}\right)
	}{\sqrt{\pi}} \mathcal{K} \left(
	-\frac{\mu}{2}, \frac{1}{2}; -\frac{ q^2 d_t^2}{2 Var}
	\right) . 
\end{align}
Where $Var$ represents the Variance of random variable $Z$, and can be expressed as $Var=(2-2q+2q^2)\sigma^2$.
$GF\left({\frac{\mu+1}{2}}\right)$ represents Gamma function as follow 
\begin{align}
	GF\left({\frac{\mu+1}{2}}\right) = \int_0^\infty x^{{\frac{\mu+1}{2}}-1} e^{-x} dx,
\end{align}
and $\mathcal{K} \left(
-\frac{\mu}{2}, \frac{1}{2}; -\frac{ q^2 d_t^2}{2 Var}
\right)$ represents Kummer's confluent hypergeometric function as
\begin{align}
	\sum_{n=0}^\infty 
	\frac{
		{
			q^{2n} d_t
		}^{2n}
	}
	{
		n! \cdot 
		4^n (1-q+q^2)^n \sigma^{2n}
	}
	\prod_{i=1}^n  \frac{
		\mu-2i+2
	}{
		1+2i-2
	} .
\end{align}

\noindent\textit{proof.} Our mathematical deduction is based on the work from \citet{winkelbauer2012moments}, and the proof is available in the appendix.

\subsection{Wasserstein Accountant in  Deep Learning}
Next, we will deduce Wasserstein accountant applied in private deep learning. We obtain Theorem 3 based on the above preparations including advanced composition, privacy loss and absolute moment under WDP. 

\noindent \textbf{Theorem 3} (Tail Bound) Under the conditions described in Theorem 2, $\mathcal{M}$ satisfies $(\mu,\varepsilon)$-WDP for 
\begin{align}
	\log \delta = \beta \sum_{t=1}^T \inf_{d_t}	\left[ \mathbb{E} \sum_{i=1}^{n}
	\left(
	\vert
	Z_{ti}
	\vert^\mu  
	\right) 
	\right]^\frac{1}{\mu}   - \beta \varepsilon .
\end{align}
Where $Z \sim	\mathcal{N}\left(q d_t,(2-2q+2q^2)\sigma^2\right)$ and $d_t = \Vert g_t-g_t'\Vert_2$. The proof of Theorem 3 is available in the appendix. In another case, if we have determined $\delta$ and want to know the privacy budget $\varepsilon$, then we can utilize the result in Corollary~1.

\noindent \textbf{Corollary 1} Under the conditions described in Theorem 2, $\mathcal{M}$ satisfies $(\mu,\varepsilon)$-WDP for 
\begin{align}
	\varepsilon= 	\sum_{t=1}^T 
	\inf_{d_t}
	\left[ \sum_{i=1}^n 
	\mathbb{E}\left(
	\vert Z_{ti} \vert^\mu
	\right) \right]^\frac{1}{\mu} 
	- \frac{1}{\beta}log\delta .
\end{align}

Corollary 1 is more commonly used than Theorem 3 since the total privacy budget generated by an algorithm plays a more important role in privacy computing.

\section{Experiments}\label{section:experiment}
The experiments in this paper consist of four parts. Firstly, we test Laplace Mechanism and Gaussian Mechanism under RDP and WDP with ever-changing orders. 
Secondly, we carry out the experiments of composition and compare our Wasserstein accountant with Bayesian accountant and moments accountant. 
Thirdly, we consider the application scenario of deep learning, and train a convolutional neural network (CNN) optimized by differentially private stochastic gradient descent (DP-SGD) \cite{DBLP:conf/ccs/AbadiCGMMT016} on the task of image classification. 
At last, we demonstrate the impact of hyperparameter variations on privacy budgets. 
All the experiments were performed on a single machine with Ubuntu 18.04, 40 Intel(R) Xeon(R) Silver 4210R CPUs @ 2.40GHz, and two NVIDIA Quadro RTX 8000 GPUs. 

\subsection{Basic Mechanisms}

\begin{figure*}[htbp]
	\centering	
	\subfigure[LM for RDP]{
		\begin{minipage}[t]{0.24\linewidth}
			\centering
			\includegraphics[width=1.6in]{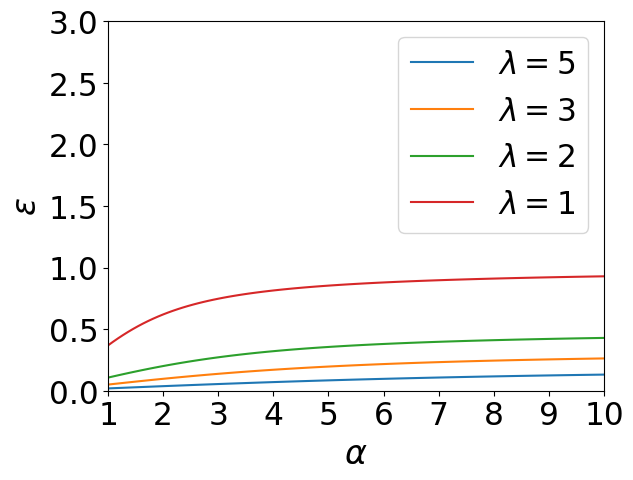}
		\end{minipage}
	}%
	\subfigure[LM for WDP]{
		\begin{minipage}[t]{0.24\linewidth}
			\centering
			\includegraphics[width=1.6in]{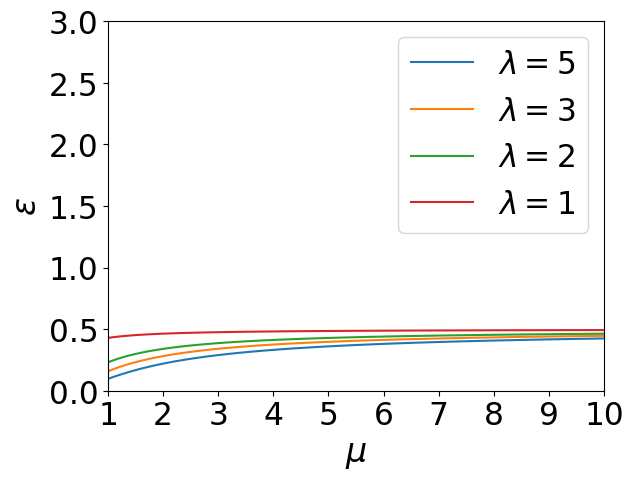}
		\end{minipage}
	}%
	\subfigure[GM for RDP]{
		\begin{minipage}[t]{0.24\linewidth}
			\centering
			\includegraphics[width=1.6in]{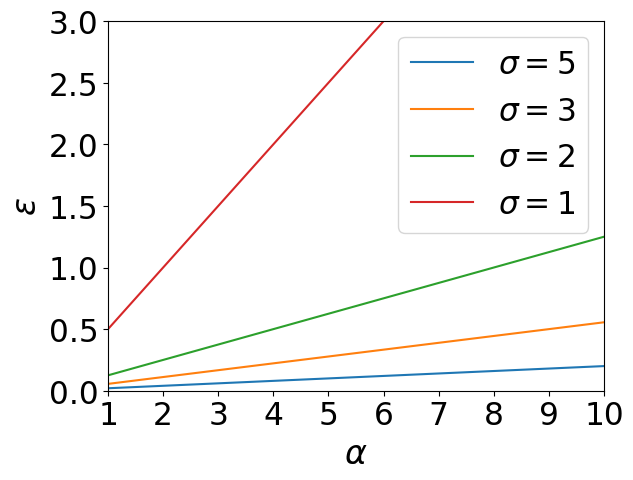}
		\end{minipage}
	}%
	\subfigure[GM for WDP]{
		\begin{minipage}[t]{0.24\linewidth}
			\centering
			\includegraphics[width=1.6in]{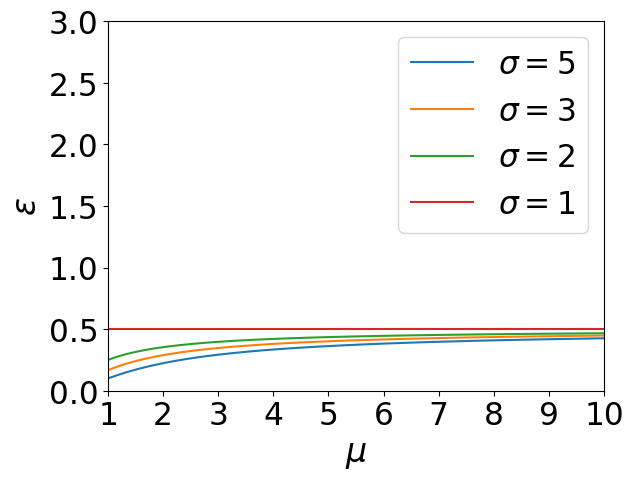}
		\end{minipage}
	}%
	\centering
	\caption{Privacy buget curves of $(\mu, \varepsilon)$-WDP and $(\alpha, \varepsilon)$-RDP  for Laplace mechanism (LM) and Gaussian mechanism (GM) with varying orders. Where $\lambda$ and $\sigma$ is the scale of LM and GM respectively. The sensitivities are set to 1 and remains unchanged.}
	\label{fig:order_epsilon_sensitivity}
\end{figure*}

We conduct experiments to test Laplace Mechanism and Gaussian Mechnism under RDP and WDP. Our experiments are based on the results of Proposition 7, 8 and Table 1. 
We set the scale parameters of Laplace mechanism and Gaussian mechanism as 1, 2, 3 and 5 respectively. The order $\mu$ of WDP is allowed to varies from 1 to 10, and so is the order $\alpha$ of RDP. We plot the values of privacy budgets $\varepsilon$ with increasing orders, and the results are shown in Figure~\ref{fig:order_epsilon_sensitivity}. 

We can observe that the privacy budgets of WDP increase with $\mu$ growing, which corresponds to our monotonicity property (see Proposition 4). More importantly, we find that the privacy budgets of WDP are not susceptible to the order $\mu$, because their curves all exhibit slow upward trends. 
However, the privacy budgets of RDP experience a steep increase under Gaussian mechanism when the noise scale equals 1, simply because its order $\alpha$ increases. In addition, the slopes of RDP curves with different noise scales are significantly different. 
These phenomena lead users to confusion about order selection and risk assessment through privacy budgets when utilizing RDP.





\subsection{Composition}\label{section:composition}

For the convenience of comparison, we adopt the same settings as the composition experiment in \citet{DBLP:conf/icml/TriastcynF20}. We imitate heavy-tailed gradient distributions by generating synthetic gradients from a Weibull distribution with $0.5$ as its shape parameter and $50 \times 1000$ as its size. 

The hyper-parameter $\sigma$ remains unchanged after being set as $0.2$, and the threshold of gradient clipping $C$ is set to $\{0.05, 0.50, 0.75, 0.99\}$-quantiles of gradient norm in turns. To observe the original variations of their privacy budgets, we do not clip gradients.  
Thus, $C$ only affects Gaussian noise with variance $C^2\sigma^2$ in DP-SGD \cite{DBLP:conf/ccs/AbadiCGMMT016} in this experiment. 
In addition, we also provide the composition results with gradient clipping in the appendix for comparison. 

In Figure \ref{figure:composition}, we have the following key observations. (1) The curves obtained from Wasserstein accountant (WA) almost replicate the changes and trends depicted by the curves obtained from moments accountant (MA) and Bayesian accountant (BA). (2) The privacy budgets under WA are always the lowest,  and this advantage becomes more significant with $C$ increasing. 



The above results show that Wasserstein accountant can retain the privacy features expressed by MA and BA at a lower privacy budget. 


\begin{figure*}[htbp]
	\centering	
	\subfigure[0.05-quantile of $\Vert g_t \Vert$]{
		\begin{minipage}[t]{0.24\linewidth}
			\centering
			\includegraphics[width=1.6in]{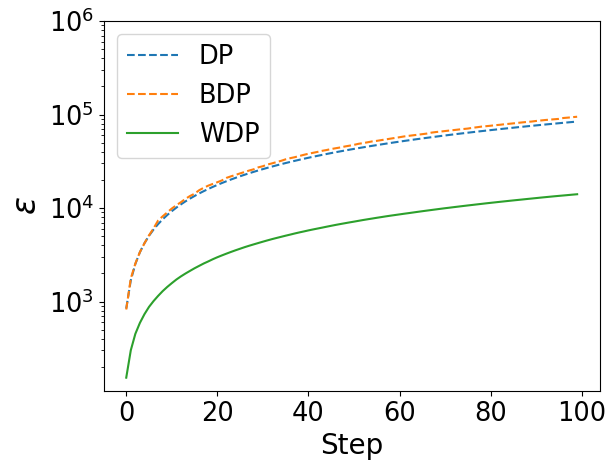}
		\end{minipage}
	}%
	\subfigure[0.50-quantile of $\Vert g_t \Vert$]{
		\begin{minipage}[t]{0.24\linewidth}
			\centering
			\includegraphics[width=1.6in]{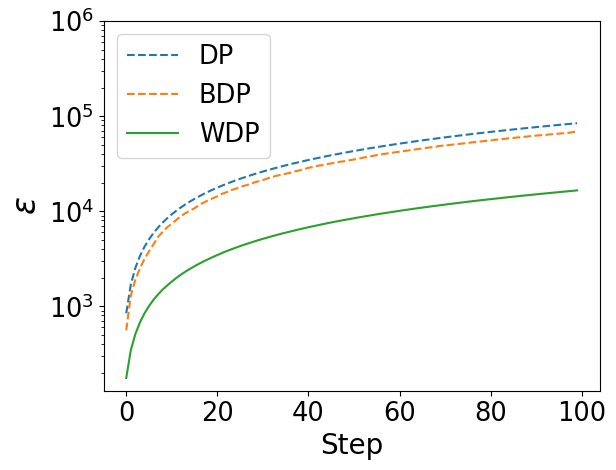}
		\end{minipage}
	}%
	\subfigure[0.75-quantile of $\Vert g_t \Vert$]{
		\begin{minipage}[t]{0.24\linewidth}
			\centering
			\includegraphics[width=1.6in]{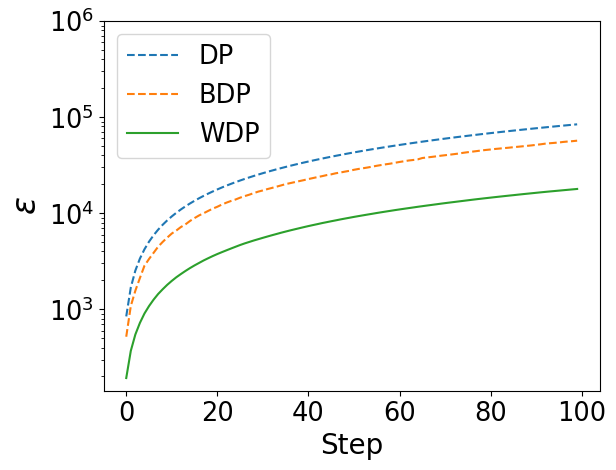}
		\end{minipage}
	}%
	\subfigure[0.99-quantile of $\Vert g_t \Vert$]{
		\begin{minipage}[t]{0.24\linewidth}
			\centering
			\includegraphics[width=1.6in]{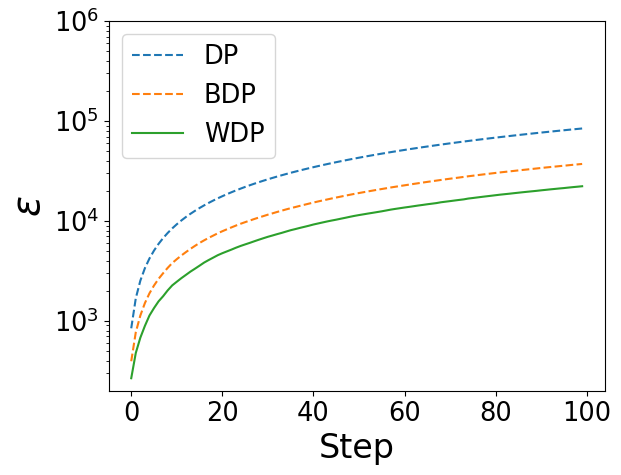}
		\end{minipage}
	}%
	\centering
	\caption{Privacy budgets over synthetic gradients obtained by moments accountant under DP, Bayesian accountant under BDP and Wasserstein accountant under WDP without gradient clipping. }
	\label{figure:composition}
\end{figure*}

\subsection{Deep Learning}

We adopt DP-SGD \cite{DBLP:conf/ccs/AbadiCGMMT016} as the private optimizer to obtain the privacy budgets under MA, BA and our WA when applying a CNN model designed by \citet{DBLP:conf/icml/TriastcynF20} to the task of image classification on four baseline datasets including MNIST \cite{726791}, CIFAR-10 \cite{2009Learning}, SVHN \cite{37648} and Fashion-MNIST \cite{xiao2017/online}. 

In the experiment of deep learning, we allow different DP frameworks to adjust the noise scale $\sigma$ according to their own needs. The reasons are as follows: (1) MA supported by DP can easily lead to gradient explosion when the noise scale is small, thus $\sigma$ can only take a relatively larger value to avoid this situation. However, an excessive noise limits the performance of BDP and WDP. (2) In addition, this setting enables our experimental results more convenient to compare with that in BDP~\citep{DBLP:conf/icml/TriastcynF20}, because the deep learning experiment in BDP is also designed in this way.

Table \ref{table:deep_learning_results} shows the results obtained under the above experimental settings. We can observe the following phenomenons: 
(1) WDP requires lower privacy budgets than DP and RDP to achieve the same level of test accuracy. 
(2) The convergence speed of the deep learning model under WA is faster than that of MA and BA. 
Taking the experiments on MNIST dataset as an example, DP and BDP need more than 100 epochs and 50 epochs of training respectively to achieve the accuracy of 96\%. While our WDP can reach the same level after only 16 epochs of training. 

BDP \cite{DBLP:conf/icml/TriastcynF20} attributes its better performance than DP to considering the gradient distribution information. 
Similarly, we can also analyze the advantages of WDP from the following aspects.  
(1) From the perspective of definition, WDP also utilizes gradient distribution information through $\gamma \in \left(Pr_\mathcal{M}(D), Pr_\mathcal{M}(D')\right)$. From the perspective of Wasserstein accountant,  the information of gradient distribution is included in $d_t$ and $Z_t$. 
(2) More importantly, privacy budgets under WDP will not explode even under low noise conditions. Because Wasserstein distance is more stable than Renyi divergence or maximum divergence, which is similar to the reason why WGAN \cite{DBLP:conf/icml/ArjovskyCB17} succeed to alleviate the problem of mode collapse by applying Wasserstein distance.

\begin{table*}[htbp]
	\begin{center}   
		\begin{tabular}{c|cc|ccc}   
			\toprule   
			&\multicolumn{2}{c|}{\textbf{Accuracy}}&\multicolumn{3}{c}{\textbf{Privacy}}\\
			\textbf{Dataset} & {Non Private} & {Private} &{DP ($\delta=10^{-5}$)}&{BDP  ($\delta=10^{-10}$)}&{WDP  ($\delta=10^{-10}$)} \\    
			\midrule   
			\multirow{1}{*}{MNIST}    & 99\%&96\%&2.2 (0.898)&0.95 (0.721)&\textbf{0.76 (0.681)}\\ 
			\multirow{1}{*}{CIFAR-10}&86\% &73\%&8.0 (0.999)&0.76 (0.681)&\textbf{0.52 (0.627)}\\   
			\multirow{1}{*}{SVHN}     &93\%&92\%&5.0 (0.999)&0.87 (0.705)&\textbf{0.40 (0.599)}\\
			\multirow{1}{*}{F-MNIST} &92\%&90\%&2.9 (0.623)&0.91 (0.713)&\textbf{0.45 (0.611)}\\
			\bottomrule 
		\end{tabular}   
	\end{center}   
	\caption{Privacy budgets accounted by DP, BDP and WDP on MNIST, CIFAR-10, SVHN and Fashion-MNIST (F-MNIST). The values in parentheses are the probability of potential attack success computed by $P(A)={1}/{(1+e^{-\varepsilon})}$ (see Section 3 in \citet{DBLP:conf/icml/TriastcynF20}).}  
	\label{table:deep_learning_results} 
\end{table*}

\subsection{Effect of $\beta$ and $\delta$}
\begin{figure}[tb]
	\centering	
	\subfigure[$\varepsilon$ varies with $\beta$]{
		\begin{minipage}[t]{0.45\linewidth}
			\centering
			\includegraphics[width=1.4in]{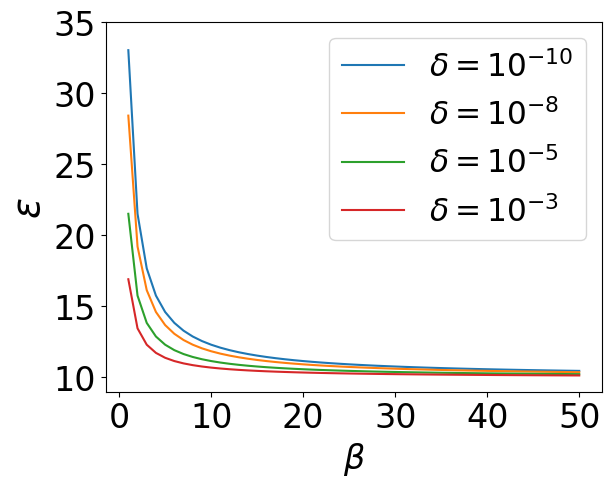}
			\label{fig:wdp_beta_epsilon}
		\end{minipage}
	}%
	\subfigure[$\varepsilon$ varies with $\delta$]{
		\begin{minipage}[t]{0.48\linewidth}
			\centering
			\includegraphics[width=1.4in]{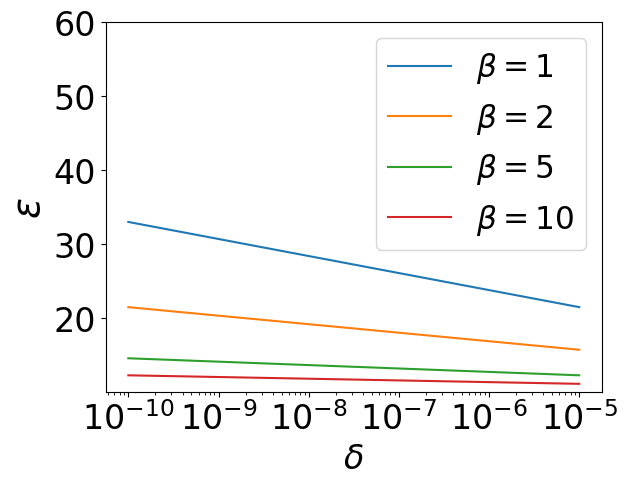}
			\label{fig:wdp_delta_epsilon}
		\end{minipage}
	}%
	\centering
	\caption{The impact of $\beta$ and $\delta$. The coordinates of horizontal axis in \ref{fig:wdp_delta_epsilon} are on a logarithmic scale.}
\end{figure}

We also conduct experiments to illustrate the relation between privacy budgets and related hyperparameters.
Our experiments are based on the results from Theorem 3 and Corollary 1, which have been proved before. 
In Figure \ref{fig:wdp_beta_epsilon}, the hyperparameter $\beta$ in WDP are allowed to varies from 1 to 50, and the failure probability $\delta$ of WDP can only be $\{10^{-10},10^{-8},10^{-5},10^{-3}\}$.
While in Figure \ref{fig:wdp_delta_epsilon}, the failure probability $\delta$ is allowed to varies from $10^{-10}$ to $10^{-5}$, and the hyperparameter $\beta$ under WDP can only be $\{1,2,5,10\}$. 
We observe that $\beta$ has a clear effect on the value of $\varepsilon$ in Figure \ref{fig:wdp_beta_epsilon}. $\varepsilon$ decreases quickly when $\beta$ is less than 10, while very slowly when it is greater than 10. When it comes to \ref{fig:wdp_delta_epsilon}, $\varepsilon$ seems to be decreasing uniformly with the exponential growth of delta.

\section{Discussion}


\subsection{Relations to Other DP Frameworks}


We establish the bridges between WDP, DP and RDP through Proposition 9, 10 and 11. We know that $\varepsilon$-DP implies $\left(\mu, \frac{1}{2}  \Delta_p f \left(  2\varepsilon \cdot  (e^\varepsilon - 1)    \right)^{\frac{1}{2\mu}}\right)$-WDP and $(\alpha, \varepsilon)$-RDP implies $\left(\mu, \frac{1}{2} \Delta_p f \left(  2\varepsilon  \right)^{\frac{1}{2 \mu}}\right)$-WDP. 
In addition, $(\mu,\varepsilon)$-WDP implies $\left(\alpha, \frac{\alpha}{\alpha-1}  L  \cdot \varepsilon^{\mu/(\mu+1)} \right)$-RDP or $\left(L  \cdot \varepsilon^{\mu/(\mu+1)}\right)$-DP.  

With the above basic conclusions, we can obtain more derivative relationships through RDP or DP. For example, we can obtain that $(\mu,\varepsilon)$-WDP implies $\frac{1}{2} \left(L  \cdot \varepsilon^{\mu/(\mu+1)} \right)^2$-zCDP (zero-concentrated differentially private) according to Proposition 1.4 in \citet{DBLP:conf/tcc/BunS16}, 

\subsection{Advantages from Metric Property}
The privacy losses of DP, RDP and BDP are all non-negative but asymmetric, and do not satisfy triangle inequality~\citep{DBLP:conf/csfw/Mironov17}. 
Several obvious advantages of WDP as a metric DP have been mentioned in the introduction (see Section \ref{section:introduction}) and verified in the experiments (see Section \ref{section:experiment}), and here we provide more additional details.


\noindent\textbf{Triangle inequality.}  
(1) Several properties including basic sequential composition, group privacy and advanced composition are derived from triangle inequality. 
(2) Properties in WDP are more comprehensible and easier to utilize than those in RDP. For example, RDP have to introduce additional conditions of $2^c$-stable and $\alpha\geq 2^{c+1}$ to derive group privacy (see Proposition 2 in \citet{DBLP:conf/csfw/Mironov17}), where $c$ is a constant.   
In contrast, our WDP utilizes its intrinsic triangle inequality to obtain group privacy without introducing any complex concepts or conditions. 

\noindent\textbf{Symmetry.} We have considered that the asymmetry of privacy loss would not be transferred to the privacy budget. Specifically, even if $D_\alpha(Pr_\mathcal{M}(D)\Vert Pr_\mathcal{M}(D'))\neq D_\alpha(Pr_\mathcal{M}(D')\Vert Pr_\mathcal{M}(D))$, $D_\alpha(Pr_\mathcal{M}(D)\Vert Pr_\mathcal{M}(D'))\leq \varepsilon$ still implies $D_\alpha(Pr_\mathcal{M}(D')\Vert Pr_\mathcal{M}(D))\leq \varepsilon$, because neighboring datasets $D$ and $D'$ can be all possible pairs. Even so, symmetrical privacy loss still has at least two advantages: (1) When computing privacy budgets, it can reduce the amount of computation for traversing adjacent datasets by half. 
(2) When proving properties, it is not necessary to exchange datasets and deduce it again like non-metric DP (e.g. see Proof of Theorem 3 in \citet{DBLP:conf/icml/TriastcynF20}). 

\subsection{Limitations} 
WDP has excellent mathematical properties as a metric DP, and can effectively alleviate exploding privacy budgets as an alternative DP framework. 
However, when the volume of data in the queried database is extremely small, WDP may release a much smaller privacy budget than other DP frameworks. Fortunately, this situation only occurs when there is very little data available in the dataset. WDP has great potential in deep learning that requires a large amount of data to train neural network models. 


\subsection{Additional Specifications}
\noindent\textbf{Other possibility.}  
Symmetry can be obtained by replacing Rényi divergence with Jensen-Shannon divergence (JSD) \cite{IEEE/TIT/1057082}. While JSD does not satisfy the triangle inequality unless we take its square root instead \cite{OSAN2018336}. 
Nevertheless, it still tends to exaggerate privacy budgets excessively, as it is defined based on divergence.

\noindent\textbf{Comparability.} Another question worth explaining is why the privacy budgets obtained by DP, RDP, and WDP can be compared. (1) Their process of computing privacy budgets follows the same mapping, namely $\mathcal{M} : \mathcal{D} \rightarrow \mathcal{R}$. (2) They are essentially measuring the differences in distributions between adjacent datasets, although their respective measurement methods are different.
(3) Privacy budgets can be uniformly transformed into the probability of successful attacks~\citep{DBLP:conf/icml/TriastcynF20}. 

\noindent\textbf{Computational problem.} Although obtaining the Wasserstein distance requires relatively high computational costs~\citep{Dudley1969TheSO, Fournier2015OnTR}, we do not need to worry about this issue. 
Because WDP does not need to directly calculate the Wasserstein distance no matter in basic privacy mechanisms or Wasserstein accountant for deep learning (see Proposition 7-8 and Theorem 1-3). 


\section{Conclusion}
In this paper, we propose an alternative DP framework called Wasserstein differential privacy (WDP) based on Wasserstein distance. 
WDP satisfies the properties of symmetry, triangle inequality and non-negativity that other DPs do not satisfy all, which enables the privacy losses under WDP to become real metrics. 
We prove that WDP has several excellent properties (see Proposition 1-13) through Lyapunov's inequality, Minkowski's inequality, Jensen's inequality, Markov's inequality, Pinsker's inequality and triangle inequality. We also derive advanced composition theorem, privacy loss and absolute moment under the postulation of WDP and finally obtain Wasserstein accountant to compute cumulative privacy budgets in deep learning (see Theorem 1-3 and Lemma 1). 
Our evaluations on basic mechanisms, compositions and deep learning show that WDP enables privacy budgets to be more stable and can effectively avoid the overestimation or even explosion on privacy.

\section{Acknowledgments}
This work is supported by National Natural Science Foundation of China (No. 72293583, No. 72293580), Science and Technology Commission of Shanghai Municipality Grant (No. 22511105901), Defense Industrial Technology Development Program (JCKY2019204A007) and Sino-German Research Network (GZ570).

\bibliographystyle{aaai24}
\bibliography{aaai24}

\onecolumn
\appendix

\section{Proof of Propositions and Theorems}

\subsection{Proof of Proposition 1}

\noindent\textbf{Proposition 1} (Symmetry). Let $\mathcal{M}$ be a $\left(\mu,\varepsilon\right)$-WDP algorithm, for any $\mu \ge 1$ and $\varepsilon \ge 0$ the following equation holds
\begin{equation}
	W_\mu\left(Pr_\mathcal{M}\left({D}\right),Pr_\mathcal{M}\left({D}^\prime\right)\right)=W_\mu\left(Pr_\mathcal{M}\left(D^\prime\right),Pr_\mathcal{M}\left(D\right)\right)\le\varepsilon. \nonumber
\end{equation}

\textit{Proof.} Considering the definition of ($\mu$,$\varepsilon$)-WDP, we have
\begin{equation}
	\begin{aligned}
		W_\mu\left(Pr_\mathcal{M}\left({D}\right),Pr_\mathcal{M}\left({D}^\prime\right)\right)
		=
		\left(\inf_{\gamma\in\Gamma\left(Pr_\mathcal{M}\left({D}\right),Pr_\mathcal{M}\left({D}^\prime\right)\right)}\int_{\mathcal{X}\times \mathcal{Y}}{{\rho\left(x,y\right)}^\mu d\gamma\left(x,y\right)}\right)^\frac{1}{\mu}\le\varepsilon. \nonumber
	\end{aligned}
\end{equation}
The symmetry of Wasserstein differential privacy is obvious for the reason that joint distribution has property $\Gamma(Pr_\mathcal{M}(D^\prime),Pr_\mathcal{M}(D))=\Gamma(Pr_\mathcal{M}(D),Pr_\mathcal{M}(D^\prime))$. 

Next, we want to proof that Kantorvich differential privacy also satisfies symmetry. 
Consider the definition of Kantorvich differential privacy, we have
\begin{equation}
	\begin{aligned}
		K\left(Pr_\mathcal{M}\left(D\right),Pr_\mathcal{M}\left(D^\prime\right)\right)
		=
		\sup_{\Vert \varphi\Vert_L\le 1}\mathbb{E}_{x\sim Pr_\mathcal{M}(D)}[\varphi(x)]-\mathbb{E}_{x\sim Pr_\mathcal{M}(D^\prime)}[\varphi(x)] 
	\end{aligned} 
\end{equation}
and
\begin{equation}
	\begin{aligned}
		K\left(Pr_\mathcal{M}\left(D^\prime\right),Pr_\mathcal{M}\left(D\right)\right)
		= 
		\sup_{\Vert \varphi\Vert_L\le 1}\mathbb{E}_{x\sim Pr_\mathcal{M}(D^\prime)}[\varphi(x)]-\mathbb{E}_{x\sim Pr_\mathcal{M}(D)}[\varphi(x)] .
	\end{aligned} 
\end{equation}
If we set $\psi\left(x\right)=-\varphi\left(x\right)$, then the above formula can be written as
\begin{equation}
	\begin{aligned}
		K\left(Pr_\mathcal{M}\left(D^\prime\right),Pr_\mathcal{M}\left(D\right)\right)
		&= \sup_{\Vert \psi\Vert_L\le 1}\mathbb{E}_{x\sim Pr_\mathcal{M}(D^\prime)}[-\psi(x)]-\mathbb{E}_{x\sim Pr_\mathcal{M}(D)}[-\psi(x)]\\
		&=\sup_{\Vert \psi\Vert_L\le 1}\mathbb{E}_{x\sim Pr_\mathcal{M}(D)}[\psi(x)]-\mathbb{E}_{x\sim Pr_\mathcal{M}(D^\prime)}[\psi(x)]
		\\&=K\left(Pr_\mathcal{M}\left(D\right),Pr_\mathcal{M}\left(D^\prime\right)\right) .
	\end{aligned} 
\end{equation}

\subsection{Proof of Proposition 2}
\noindent \textbf{Proposition 2}  (Triangle Inequality) Let $D_1,D_2,D_3\in \mathcal{D}$ be three arbitrary datasets. Suppose there are fewer different data entries between $D_1$ and $D_2$ compared with $D_1$ and $D_3$, and the differences between $D_1$ and $D_2$ are included in the differences between $D_1$ and $D_3$. For any randomized algorithm $\mathcal{M}$ satisfies $(\mu, \varepsilon)$-WDP with $\mu \ge 1$, we have
\begin{equation}
	\begin{aligned}
		W_{\mu}(Pr_\mathcal{M}(D_1),Pr_\mathcal{M}(D_3))
		\le W_{\mu}(Pr_\mathcal{M}(D_1),Pr_\mathcal{M}(D_2))+W_{\mu}(Pr_\mathcal{M}(D_2),Pr_\mathcal{M}(D_3)) .
	\end{aligned}
\end{equation}

\textit{Proof.} Triangle inequality has been proved by Proposition 2.1 in \citet{Philippe2008Elementary}. Here we provide a simpler proof method from another perspective. 

Firstly, we introduce another mathematical form that defines the Wasserstein distance (see Definition 6.1 in \citet{Ludger2009Optimal} or Equation 1 in \citet{Panaretos2019Statistical})
\begin{equation}
	W_\mu\left(P,Q\right) = \inf_{X\sim P\atop Y\sim Q}   \left[\mathbb{E}\; \rho(X,Y)^\mu\right]^{\frac{1}{\mu}}, \mu\geq 1 . \label{equation:wasserstien_distance_definition:expectation_form}
\end{equation}
Where $X$ and $Y$ are random vectors, and the infimum is taken over all possible pairs of $X$ and $Y$ that are marginally distributed as $P$ and $Q$.  

Let $X_1,X_2,X_3$ be three random variables follow distributions $Pr_\mathcal{M}(D_1),Pr_\mathcal{M}(D_2),Pr_\mathcal{M}(D_3)$ respectively. 
\begin{align}
	W_\mu(Pr_\mathcal{M}(D_1),Pr_\mathcal{M}(D_3))&=
	\inf_{X_1\sim Pr_\mathcal{M}(D_1)\atop X_3\sim Pr_\mathcal{M}(D_3)}   \left[\mathbb{E}\; \rho\left( X_1,X_3\right)^{\mu}\right]^{\frac{1}{\mu}}
	\\
	&\leq \inf_{X_1\sim Pr_\mathcal{M}(D_1)\atop X_2\sim Pr_\mathcal{M}(D_2)}   \left[\mathbb{E} \;\rho\left(X_1,X_2\right)^{\mu}\right]^{\frac{1}{\mu}} 
	+
	\inf_{X_2\sim Pr_\mathcal{M}(D_2)\atop X_3\sim Pr_\mathcal{M}(D_3)}   \left[\mathbb{E}\; \rho\left(X_2,X_3\right)^{\mu}\right]^{\frac{1}{\mu}} \label{equation:triangle_inequality:Minkowski_inequality}
	\\&=W_{\mu}(Pr_\mathcal{M}(D_1),Pr_\mathcal{M}(D_2))+W_{\mu}(Pr_\mathcal{M}(D_2),Pr_\mathcal{M}(D_3)).
\end{align}
Here Equation \ref{equation:triangle_inequality:Minkowski_inequality} can be established by applying Minkowski's inequality that $\Vert X_1 + X_2\Vert_r \le \Vert X_1 \Vert_r + \Vert X_2 \Vert_r$ with $1<r<\infty$. 

\subsection{Proof of Proposition 3}
\noindent\textbf{Proposition 3} (Non-Negativity). For $\mu\ge 1$ and any randomized algorithm $\mathcal{M}$, we have $W_{\mu}(Pr_\mathcal{M}(D), Pr_\mathcal{M}(D'))\ge 0$. 

\noindent\textit{Proof.}  We can be sure that the integrand function $\rho(x,y)\ge 0$, for the reason that it's a cost function in the sense of optimal transport \cite{Ludger2009Optimal} and a norm in the statistical sense \cite{Panaretos2019Statistical}. $\gamma(x,y)$ is the probability measure, so that $\gamma(x,y)>0$ holds. Then according to the definition of WDP, the integral function
\begin{equation}
	\left(\inf_{\gamma\in\Gamma\left(Pr_\mathcal{M}\left({D}\right), Pr_\mathcal{M}\left({D}^\prime\right)\right)}\int_{\mathcal{X}\times \mathcal{Y}}{{\rho\left(x,y\right)}^\mu d\gamma\left(x,y\right)}\right)^\frac{1}{\mu}\ge 0.\nonumber
\end{equation}

\subsection{Proof of Proposition 4}
\noindent \textbf{Proposition 4} (Monotonicity). For $1 \le \mu_1\le \mu_2$, we have $W_{\mu_1}(Pr_\mathcal{M}(D),Pr_\mathcal{M}(D'))\le W_{\mu_2}(Pr_\mathcal{M}(D),Pr_\mathcal{M}(D'))$, or we can equivalently described this proposition as $(\mu_2,\varepsilon)$-WDP implies $(\mu_{1},\varepsilon)$-WDP. 

\textit{Proof.} Consider the expectation form of Wasserstein differential privacy (see Equation \ref{equation:wasserstien_distance_definition:expectation_form}), and apply Lyapunov's inequality as follow
\begin{equation}
	\left[ \mathbb{E} \vert \cdot \vert^{\mu_1} \right]^{\frac{1}{\mu_1}} \le [ \mathbb{E} \vert \cdot \vert ^{\mu_{2}}]^{\frac{1}{\mu_{2}}}, 1\le \mu_{1} \le \mu_{2}
\end{equation}
we obtain that 
\begin{equation}
	\begin{aligned}
		W_{\mu_1}(Pr_\mathcal{M}(D),Pr_\mathcal{M}(D^\prime)) &=  \inf_{X\sim \mathcal{M}(D)\atop Y\sim \mathcal{M}(D^\prime)}   \left[\mathbb{E}\; \rho\left(X,Y\right)^{\mu_{1}}\right]^{\frac{1}{\mu_{1}}} 
		\\
		&\le \inf_{X\sim \mathcal{M}(D)\atop Y\sim \mathcal{M}(D^\prime)}   \left[\mathbb{E}\; \rho \left(X,Y\right)^{\mu_{2}}\right]^{\frac{1}{\mu_{2}}} 
		\\
		&= W_{\mu_2}(Pr_\mathcal{M}(D),Pr_\mathcal{M}(D^\prime)).
	\end{aligned}
\end{equation}

\subsection{Proof of Proposition 5}
\noindent\textbf{Proposition 5} (Parallel Composition). Suppose a dataset $D$ is divided into $n$ parts disjointly which are denoted as $D_{i}, i=1,2,\cdots,n$. Each randomized algorithm $\mathcal{M}_i$  performed on different seperated dataset $D_i$ respectively. If $\mathcal{M}_i:\mathcal{D}\rightarrow\mathcal{R}_i$ satisfies $\left(\mu,\varepsilon_i\right)$-WDP for $i=1,2,\cdots,n$, then a set of  randomized algorithms  $\mathcal{M}=\{\mathcal{M}_1,\mathcal{M}_2,\cdots,\mathcal{M}_n\}$ satisfies ($\mu$, $\max\{\varepsilon_1,\varepsilon_2,\cdots,\varepsilon_n\}$)-WDP.

\textit{Proof.} 
From the definition of WDP, we obtain that

\begin{align}
	W_{\mu}\left(Pr_\mathcal{M}\left(D^\prime\right),Pr_\mathcal{M}\left(D\right)\right)
	&=
	\left(\inf_{\gamma\in\Gamma\left(Pr_\mathcal{M}\left(D\right),Pr_\mathcal{M}\left(D^\prime\right)\right)}\int_{\mathcal{X}\times \mathcal{Y}}{{\rho\left(x,y\right)}^\mu d\gamma\left(x,y\right)}\right)^\frac{1}{\mu}
	\\
	\le&
	\max
	\left\{\left(\inf_{\gamma\in\Gamma\left(Pr_{\mathcal{M}_i}\left(D_i\right),Pr_{\mathcal{M}_i}\left(D_i^\prime\right)\right)}\int_{\mathcal{X}\times \mathcal{Y}}{{\rho\left(x,y\right)}^\mu d\gamma\left(x,y\right)}\right)^\frac{1}{\mu}, 
	\forall \mathcal{M}_i \subseteq \mathcal{M},D_i \subseteq D\right\}    \label{equation:parallel_composition_proof:marginal_distribution}
	\\\le& \max\{\varepsilon_1,\varepsilon_2,\cdots,\varepsilon_n\}.
\end{align}
Inequality \ref{equation:parallel_composition_proof:marginal_distribution} is tenable for the following reasons. (1) Privacy budget in WDP framework focuses on the upper bound of privacy loss or distance. (2) The randomized algorithm in $\mathcal{M}$ that leads to the maximum differential privacy budget is a certain $\mathcal{M}_i$, because only one differential privacy mechanism can be applied in both $W_{\mu}\left(Pr_{\mathcal{M}}\left(D^\prime\right),Pr_{\mathcal{M}}\left(D\right)\right)$ and $W_\mu (Pr_{\mathcal{M}_i}(D_i),Pr_{\mathcal{M}_i}(D_i))$. (3) There is only one element difference between both $D$, $D'$ and $D_i$, $D'_i$, the difference is greater when the data volume is small from the perspective of entire distributions. The query algorithm in differential privacy requires hiding individual differences, and a larger amount of data helps to hide individual data differences.


\subsection{Proof of Proposition 6}
\noindent \textbf{Proposition 6} (Sequential Composition). Consider a series of randomized algorithms $\mathcal{M}=\{\mathcal{M}_1,\cdots, \mathcal{M}_i, \cdots, \mathcal{M}_n\}$  performed on a dataset sequentially. If any $\mathcal{M}_i:\mathcal{D}\rightarrow\mathcal{R}_i$ satisfies $\left(\mu,\varepsilon_i\right)$-WDP, then $\mathcal{M}$ satisfies $(\mu, \sum_{i=1}^{n}\varepsilon_i)$-WDP.

\textit{Proof.} Consider the mathematical forms of $\left(\mu,\varepsilon_i\right)$-WDP
\begin{equation}
	\begin{aligned}
		\begin{cases}
			W_\mu(Pr_{\mathcal{M}_{1}}(D), Pr_{\mathcal{M}_{1}}(D^\prime)) \le \varepsilon_1,\\
			W_\mu(Pr_{\mathcal{M}_{2}}(D), Pr_{\mathcal{M}_{2}}(D^\prime)) \le \varepsilon_2,\\
			\qquad\cdots\\
			W_\mu (Pr_{\mathcal{M}_n}(D), Pr_{\mathcal{M}_n}(D'))\le \varepsilon_n
		\end{cases}
	\end{aligned}
\end{equation}
According to the basic properties of the inequality, we can obtain the upper bound of the sum of Wassestein distances
\begin{equation}
	\begin{aligned}
		\sum_{i=1}^{n} W_\mu(Pr_{\mathcal{M}_{i}}(D),Pr_{\mathcal{M}_{i}}(D^\prime)) \le \sum_{i=1}^{n}\varepsilon_i .
	\end{aligned}
\end{equation}
According to the triangle inequality of Wasserstein distance (see Proposition 2), we have
\begin{equation}
	\begin{aligned}
		\sum_{i=1}^{n} W_\mu(Pr_{\mathcal{M}_{i}}(D), Pr_{\mathcal{M}_{i}}(D^\prime))\ge W_\mu(Pr_{\mathcal{M}}(D),Pr_{\mathcal{M}}(D^\prime)) . 
	\end{aligned}
\end{equation}
Thus, we obtain that $W_\mu(Pr_{\mathcal{M}}(D),Pr_{\mathcal{M}}(D^\prime))\le\sum_{i=1}^{n}\varepsilon_i$.

\subsection{Proof of Proposition 7}
\noindent\textbf{Proposition 7} (Laplace Mechanism). If an algorithm $f:\mathcal{D}\rightarrow\mathcal{R}$ has sensitivity $\Delta_p f$ and the order $\mu\geq1$, then the Laplace mechanism $\mathcal{M}_L=f\left(x\right)+Lap\left(0,\lambda\right)$  preserves $\left(\mu, \frac{1}{2} \Delta_p f \left(\sqrt{2\left[ 1/\lambda + exp(-1/\lambda)-1 \right]} \right)^{\frac{1}{\mu}} \right)$-Wasserstein differential privacy. 

\textit{Proof.}  Considering the Wasserstein distance between two Laplace distributions, we have 
\begin{align}
	W_\mu\left(Lap\left(0,\lambda\right), Lap\left(\Delta_p f,\lambda\right)\right)\label{equation:start_Laplace_mechanism}
	&=\left(\inf_{\gamma\in\Gamma\left(Lap\left(0,\lambda\right), Lap\left(\Delta_p f,\lambda\right)\right)}\int_{\mathcal{X}\times \mathcal{Y}}{{\rho\left(x,y\right)}^\mu d\gamma\left(x,y\right)}\right)^\frac{1}{\mu}
	\\
	&\leq \left(\inf_{\gamma\in\Gamma\left(Lap\left(0,\lambda\right), Lap\left(\Delta_p f,\lambda\right)\right)}\int_{\mathcal{X}\times \mathcal{Y}}{{ \Delta_p f }^\mu d\gamma\left(x,y\right)}\right)^\frac{1}{\mu}
	\\
	& = \Delta_p f \left(\inf_{\gamma\in\Gamma\left(Lap\left(0,\lambda\right), Lap\left(\Delta_p f,\lambda\right)\right)} \int  1 d  \gamma\left(x,y\right)     \right)^{\frac{1}{\mu}}
	\\
	& = \Delta_p f  \inf_{X\sim {Lap}\left(0,\lambda\right) \atop Y\sim {Lap} \left(\Delta_p f,\lambda\right)}  \left[\mathbb{E}\;1_{X\not = Y}\right]^{\frac{1}{\mu}} \label{equation:opt_inequality_pre} 
	\\
	& = \frac{1}{2} \Delta_p f \left(  \Vert {Lap}\left(0,\lambda\right) - Lap(\Delta_p f,\lambda) \Vert_{TV} \right)^{\frac{1}{\mu}}  \label{equation:opt_inequality}
	\\
	& \leq \frac{1}{2}  \Delta_p f  \left( \sqrt{2 D_{KL}({Lap}\left(0,\lambda\right)
		\Vert 
		{Lap}\left(\Delta_p f,\lambda\right)
		)}  \right)^{\frac{1}{\mu}}. \label{equation:Pinsker_inequality}
\end{align}
Where $\Delta_p f$ is the $l_p$-sensitivity between two datasets (see Definition 8), and $p$ is its order which can be set to any positive integer as needed. $X$ and $Y$ are random variables follows Laplace distribution (see Equation \ref{equation:opt_inequality_pre}).
In addition, $\Vert  \cdot \Vert_{TV}$ represents the total variation. $D_{KL}(P\Vert Q)$ represents the Kullback–Leibler (KL) divergence between $P$ and $Q$, which is also equal to one-order Rényi divergence $D_1(P\Vert Q)$ (see Theorem 5 in \citet{DBLP:journals/tit/ErvenH14} or Definition 3 in \citet{DBLP:conf/csfw/Mironov17}).

We can obtain Equation \ref{equation:opt_inequality} from Equation \ref{equation:opt_inequality_pre}  because of the probabilistic interpretation of total variation when $\rho(x,y)=1$, which has been proposed at page 10 in Reference \cite{Ludger2009Optimal}. 
Equation \ref{equation:Pinsker_inequality} can be established because of 
Pinsker's inequality (see Section I in \citet{DBLP:journals/tit/FedotovHT03})
\begin{equation}
	D_{KL}(P\Vert Q)\geq \frac{1}{2} \Vert P-Q \Vert_{TV}^2.
\end{equation}
Pinsker's inequality establishs a relation between KL divergence and total variation, and $P$ and $Q$ represent the distributions of two random variables respectively, and

To obtain the final result, we 
apply the outcome of Laplace Mechanism under Rényi DP of order one (see Table II in  \citet{DBLP:conf/csfw/Mironov17}) as follow
\begin{equation}
	D_1(Lap(0,\lambda)\Vert Lap(1,\lambda))	= 1/\lambda + exp(-1/\lambda)-1.
\end{equation}
Then we will obtain the outcome of Laplace Mechnism under wasserstein DP as follow
\begin{align}
	W_\mu\left(Lap\left(0,\lambda\right), Lap\left(1,\lambda\right)\right) 
	\leq  
	\frac{1}{2}\Delta_p f\left( \sqrt{2\left[ 1/\lambda + exp(-1/\lambda)-1 \right]} \right)^{\frac{1}{\mu}}.
\end{align}

\subsection{Proof of Proposition 8}
\noindent\textbf{Proposition 8} (Gaussian Mechanism). If an algorithm $f:\mathcal{D}\rightarrow\mathcal{R}$ has sensitivity $\Delta_p f$ and the order $\mu\geq 1$,  then Gaussian mechanism $\mathcal{M}_G=f\left(x\right)+\mathcal{N}\left(0,\sigma^2\right)$ preserves
$\left(\mu, \frac{1}{2} \left({\Delta_p f}/{\sigma} \right)^{\frac{1}{\mu}} \right)$-Wasserstein differential privacy.

\textit{Proof.} By directly calculating the Wasserstein distance between Gaussian distributions, we have
\begin{align}
	W_\mu\left(\mathcal{N}\left(0,\sigma^2\right), \mathcal{N}\left(\Delta_p f,\sigma^2\right)\right)
	&=\left(\inf_{\gamma\in\Gamma \left(\mathcal{N}\left(0,\sigma^2\right), \mathcal{N}\left(\Delta_p f,\sigma^2\right)\right)}\int_{\mathcal{X}\times \mathcal{Y}}{{\rho\left(x,y\right)}^\mu d\gamma\left(x,y\right)}\right)^\frac{1}{\mu}
	\label{equation:start_Gaussian_mechanism}
	\\
	&\leq \left(\inf_{\gamma\in\Gamma\left(\mathcal{N}\left(0,\sigma^2\right), \mathcal{N}\left(\Delta_p f,\sigma^2\right)\right)}\int_{\mathcal{X}\times \mathcal{Y}}{{ \Delta_p f }^\mu d\gamma\left(x,y\right)}\right)^\frac{1}{\mu}
	\\
	& = \Delta_p f \left(\inf_{\gamma\in\Gamma\left(\mathcal{N}\left(0,\sigma^2\right), \mathcal{N}\left(\Delta_p f,\sigma^2\right)\right)} \int  1 d  \gamma\left(x,y\right)     \right)^{\frac{1}{\mu}}
	\\
	& = \Delta_p f  \inf_{X\sim {\mathcal{N}}\left(0,\sigma^2\right) \atop Y\sim {\mathcal{N}} \left(\Delta_p f,\sigma^2\right)}  \left[\mathbb{E}\;1_{X\not = Y}\right]^{\frac{1}{\mu}} \label{equation:opt_inequality_pre_Gaussian} 
	\\
	& = \frac{1}{2} \Delta_p f \left(  \Vert \mathcal{N}(0, \sigma^2)- \mathcal{N}(\Delta_p f, \sigma^2) \Vert_{TV} \right)^{\frac{1}{\mu}}  \label{equation:opt_inequality_Gaussian}
	\\
	& \leq \frac{1}{2}  \Delta_p f  \left( \sqrt{2 D_{KL}\left(\mathcal{N}(0, \sigma^2)\Vert \mathcal{N}(\Delta_p f, \sigma^2)\right)}  \right)^{\frac{1}{\mu}}. \label{equation:Pinsker_inequality_Gaussian}
\end{align}
Where $\Delta_p f$ is the $l_p$-sensitivity between two datasets (see Definition 8). $X$ and $Y$ are random variables follows Gaussian distribution. $\Vert  \cdot \Vert_{TV}$ represents the total variation. $D_{KL}(P\Vert Q)$ represents the KL divergence between $P$ and $Q$, which is also equal to one-order Rényi divergence $D_1(P\Vert Q)$ (see Theorem 5 in \citet{DBLP:journals/tit/ErvenH14} or Definition 3 in \citet{DBLP:conf/csfw/Mironov17}).

We can obtain Equation \ref{equation:opt_inequality_Gaussian} from Equation \ref{equation:opt_inequality_pre_Gaussian}  because of the probabilistic interpretation of total variation when $\rho(x,y)=1$ (see page 10 in \citet{Ludger2009Optimal}). 
Equation \ref{equation:Pinsker_inequality_Gaussian} can be established because of 
Pinsker's inequality (see Section I in \citet{DBLP:journals/tit/FedotovHT03})
\begin{equation}
	D_{KL}(P\Vert Q)\geq \frac{1}{2} \Vert P-Q \Vert_{TV}^2.
\end{equation}
Pinsker's inequality establishs a relation between KL divergence and total variation, and $P$ and $Q$ represent the distributions of two random variables. 

To obtain the final result, we 
apply the property of Gaussian Mechanism under Rényi DP of order one (see Proposition 7 and Table II in \citet{DBLP:conf/csfw/Mironov17}) as follow
\begin{equation}
	D_1(\mathcal{N}(0,\sigma^2)\Vert \mathcal{N}(1,\sigma^2))	= \frac{(\Delta_p f)^2}{2 \sigma^2}.
\end{equation}
Then we will obtain the outcome of Gaussian Mechnism under wasserstein DP as follow
\begin{equation}
	\begin{aligned}
		W_\mu\left(\mathcal{N}\left(0,\sigma^2\right), \mathcal{N}\left(1,\sigma^2\right)\right) 
		\leq  
		\frac{1}{2}\left( \sqrt{2 \frac{(\Delta_p f)^2}{2 \sigma^2}} \right)^{\frac{1}{\mu}}
		=\frac{1}{2} \left(\frac{\Delta_p f}{\sigma} \right)^{\frac{1}{\mu}}. 
	\end{aligned}
\end{equation}
Thus we have proved that if algorithm $f$ has sensitivity 1, then the Gaussian mechanism $\mathcal{M}_G$ satisfies $\left(\mu, \frac{1}{2} \left({\Delta_p f}/{\sigma} \right)^{\frac{1}{\mu}} \right)$-WDP.


\subsection{Proof of Proposition 9}

\noindent\textbf{Proposition 9} (From DP to WDP) If $\mathcal{M}$ preserves $\varepsilon$-DP with sensitivity $\Delta f$
, it also satisfies $\left(\mu, \frac{1}{2} \Delta_p f \left({2 \varepsilon \cdot (e^\varepsilon-1) }\right)^{\frac{1}{2 \mu}} \right)$-WDP.

\textit{Proof.} Considering the definition of Wasserstein differential privacy and refering to Equation \ref{equation:start_Laplace_mechanism}-\ref{equation:Pinsker_inequality}, we have
\begin{align}
	W_\mu\left(Pr_\mathcal{M}(D),Pr_\mathcal{M}(D')\right)
	\leq 
	\frac{1}{2}  \Delta_p f  \left( \sqrt{2 D_{KL} (Pr_\mathcal{M}(D) \Vert Pr_\mathcal{M}(D'))}  \right)^{\frac{1}{\mu}}. \label{equation:RDP2WDP}
\end{align}

To deduce further, we apply Lemma 3.18 in \citet{DBLP:journals/fttcs/DworkR14}. It said that if two random variables $X$, $Y$ satisfy $D_{\infty}(X\Vert Y)\leq \varepsilon$ and $D_{\infty}(X\Vert Y)\leq\varepsilon$, then we can obtain 
\begin{align}
	D_{1}(X\Vert Y)\leq \varepsilon \cdot (e^\varepsilon-1). \label{equation:Lemma3.18}
\end{align}

It should be noted that the condition of $\varepsilon$-DP ensures
that $D_{\infty}(X\Vert Y)\leq \varepsilon$ and $D_{\infty}(X\Vert Y)\leq\varepsilon$ can be established (see Remark 3.2 in \cite{DBLP:journals/fttcs/DworkR14}).
Based on Equation \ref{equation:RDP2WDP} and Equation \ref{equation:Lemma3.18}, we have
\begin{align}
	W_\mu(Pr_\mathcal{M}(D),Pr_\mathcal{M}(D')) \leq \frac{1}{2} \Delta_p f \left(\sqrt{2 \varepsilon \cdot (e^\varepsilon-1) }\right)^{\frac{1}{\mu}} 
	=
	\frac{1}{2} \Delta_p f \left({2 \varepsilon \cdot (e^\varepsilon-1) }\right)^{\frac{1}{2 \mu}} 
	.
\end{align}

\subsection{Proof of Proposition 10}

\noindent \textbf{Proposition 10} (From RDP to WDP) If $\mathcal{M}$ preserves $(\alpha,\varepsilon)$-RDP with sensitivity $\Delta_p f$, it also satisfies $\left(\mu,\frac{1}{2} \Delta_p f  \left(2\varepsilon\right)^{\frac{1}{2\mu}} \right)$-WDP.

\textit{Proof.} Considering the definition of Wasserstein differential privacy and refering to Equation \ref{equation:start_Laplace_mechanism}-\ref{equation:Pinsker_inequality}, we have
\begin{equation}\label{equation:RDP2WDP_origin}
	\begin{aligned}
		W_\mu\left(Pr_\mathcal{M}(D),Pr_\mathcal{M}(D')\right) 
		\leq 
		\frac{1}{2}  \Delta_p f  \left( \sqrt{2 D_{KL}(Pr_\mathcal{M}(D)\Vert Pr_\mathcal{M}(D'))}  \right)^{\frac{1}{\mu}}.
	\end{aligned}
\end{equation}
Where $D_{KL} (Pr_\mathcal{M}(D)\Vert Pr_\mathcal{M}(D'))$ represents the KL divergence between $Pr_\mathcal{M}(D)$ and $Pr_\mathcal{M}(D')$, which can also written as 1-order Rényi divergence (see Theorem 5 in \citet{DBLP:journals/tit/ErvenH14} or Definition 3 in \citet{DBLP:conf/csfw/Mironov17})
\begin{align}\label{equation:KL_in_proposition10}
	D_{KL} (Pr_\mathcal{M}(D)\Vert Pr_\mathcal{M}(D')) = D_1(Pr_\mathcal{M}(D), Pr_\mathcal{M}(D')) .
\end{align} 
In addition, from the monotonicity property of RDP, we have
\begin{align}\label{equation:monocity_of_RDP}
	D_{\mu_1}(Pr_\mathcal{M}(D),Pr_\mathcal{M}(D')) \leq D_{\mu_2}(Pr_\mathcal{M}(D),Pr_\mathcal{M}(D'))
\end{align}
for $1 \leq \mu_1 < \mu_2$ and arbitrary $Pr_\mathcal{M}(D)$ and $Pr_\mathcal{M}(D')$.

From the condition that $\mathcal{M}$ preserves $(\alpha,\varepsilon)$-RDP, we have 
\begin{align}\label{equation:RDP_in_proposition10}
	D_\alpha (Pr_\mathcal{M}(D),Pr_\mathcal{M}(D')) \leq \varepsilon,\;  \alpha\geq 1
\end{align}
Combining Equation \ref{equation:KL_in_proposition10}, \ref{equation:monocity_of_RDP} and \ref{equation:RDP_in_proposition10}, we have
\begin{align}\label{equation:combinition}
	D_{KL} (Pr_\mathcal{M}(D)\Vert Pr_\mathcal{M}(D')) = 
	D_1 (Pr_\mathcal{M}(D), Pr_\mathcal{M}(D')) 
	\leq
	D_\alpha (Pr_\mathcal{M}(D), Pr_\mathcal{M}(D')) \leq \varepsilon .
\end{align}
Combining Equation \ref{equation:RDP2WDP_origin} and \ref{equation:combinition}, we have
\begin{align}\label{equation:1-rdp2wdp}
	W_\mu\left(Pr_\mathcal{M}(D), Pr_\mathcal{M}(D')\right) 
	\leq 
	\frac{1}{2} \Delta_p f \left( \sqrt{2 \varepsilon}  \right)^{\frac{1}{\mu}} 
	= 
	\frac{1}{2} \Delta_p f \left(2\varepsilon\right)^{\frac{1}{2\mu}}.
\end{align}

Therefore, $(\alpha,\varepsilon)$-RDP implies $\left(\mu,\frac{1}{2} \Delta_p f \left(2\varepsilon\right)^{\frac{1}{2\mu}} \right)$-WDP.

\subsection{Proof of Proposition 11}
\noindent \textbf{Proposition 11} (From WDP to RDP) Suppose $\mu \geq 1$ and $\log(p_\mathcal{M}(\cdot))$ is an $L$-Lipschitz function. 
If $\mathcal{M}$ preserves $(\mu,\varepsilon)$-WDP with sensitivity $\Delta_p f$, it also satisfies $\left(\alpha, \frac{\alpha}{\alpha-1}  L  \cdot \varepsilon^{\mu/(\mu+1)} \right)$-RDP.
Specifically, when $\alpha \rightarrow \infty$, it satisfies $\left(L  \cdot \varepsilon^{\mu/(\mu+1)}\right)$-DP. 

\noindent\textit{proof.} Considering the definition of $L$-Lipschitz function, we have
\begin{align}
	\vert \log p_\mathcal{M}(D) - \log p_\mathcal{M}(D') \vert &\leq L \vert p_\mathcal{M}(D) -  p_\mathcal{M}(D') \vert
	\\
	\Bigg\vert \log \frac{p_\mathcal{M}(D)}{p_\mathcal{M}(D')} \Bigg\vert &\leq L \vert p_\mathcal{M}(D)  - p_\mathcal{M}(D') \vert 
	\\
	-L \vert p_\mathcal{M}(D) - p_\mathcal{M}(D') \vert
	&\leq
	\log \frac{p_\mathcal{M}(D)}{p_\mathcal{M}(D')}  
	\leq 
	L \vert p_\mathcal{M}(D) - p_\mathcal{M}(D') \vert
	\\
	e^{-L \vert p_\mathcal{M}(D) - p_\mathcal{M}(D') \vert}  \label{equation:wdp2rdp_pre_ine_log}
	&\leq 
	\frac{p_\mathcal{M}(D)}{p_\mathcal{M}(D')} 
	\leq 
	e^{L \vert p_\mathcal{M}(D) - p_\mathcal{M}(D') \vert}.
\end{align}

Considering the Rényi divergence with order $\alpha$, we have
\begin{align}
	D_\alpha (Pr_{\mathcal{M}}(D) \Vert Pr_{\mathcal{M}}(D')) 
	&= \frac{1}{\alpha-1} \log 
	\mathbb{E}_{Pr_{\mathcal{M}}(D')} \left[ \left(
	\frac{p_{\mathcal{M}}(D)}{p_{\mathcal{M}}(D')}
	\right)^\alpha \right]
	\\
	&\leq
	\frac{1}{\alpha-1} \log 
	\mathbb{E}_{Pr_{\mathcal{M}}(D')} 
	\left(
	e^{\alpha L \vert p_{\mathcal{M}}(D) - p_\mathcal{M}(D') \vert}
	\right) 
	\\
	&\leq 
	\frac{1}{\alpha-1} \log \mathbb{E}_{Pr_\mathcal{M}(D')} \left(
	e^{\alpha L \Delta_p f}
	\right) \label{equation:wdp2rdp_sensitivity}.
\end{align}

According to the definition of sensitivity, we know that
\begin{align}
	\begin{cases}
		p_{\mathcal{M}}(D) \leq p_{\mathcal{M}}(D') + \Delta_p f,  &p_\mathcal{M}(D) \geq p_\mathcal{M}(D')
		\\
		p_{\mathcal{M}}(D')\leq p_{\mathcal{M}}(D) + \Delta_p f,  &p_\mathcal{M}(D) \leq p_\mathcal{M}(D')
	\end{cases} \label{equation:wdp2rdp_inequality} .
\end{align}

From Theorem 2.7 in \citet{2019One}, we have
\begin{align}
	\Delta_p f \leq W_\mu (Pr_{\mathcal{M}}(D) \Vert Pr_{\mathcal{M}}(D'))^{\mu/(\mu+1)}   \label{equation:wdp2rdp_most_important} .
\end{align}

Combining Equation \ref{equation:wdp2rdp_sensitivity} and \ref{equation:wdp2rdp_most_important}, we have
\begin{align}
	D_\alpha (Pr_{\mathcal{M}}(D) \Vert Pr_{\mathcal{M}}(D')) 
	&\leq
	\frac{1}{\alpha-1} \log \mathbb{E}_{Pr_\mathcal{M}(D')} \left(
	e^{\alpha L [W_\mu (Pr_{\mathcal{M}}(D) \Vert Pr_{\mathcal{M}}(D'))]^{\mu/(\mu+1)}}
	\right) 
	\\
	&=
	\frac{1}{\alpha-1}\log 
	\mathbb{E}_{Pr_\mathcal{M}(D')}
	\left(
	e^{\alpha L \varepsilon^{\mu/(\mu+1)}}
	\right) 
	\\
	&=
	\frac{1}{\alpha-1}\log 
	\left(
	e^{\alpha L \varepsilon^{\mu/(\mu+1)}}
	\right)
	\\&=
	\frac{\alpha}{\alpha-1}  L \varepsilon^{\mu/(\mu+1)}. 
\end{align}


Through the same methods, we can also prove that 
\begin{align}
	D_\alpha (Pr_{\mathcal{M}}(D') \Vert Pr_{\mathcal{M}}(D)) \leq \frac{\alpha}{\alpha-1}  L  \varepsilon^{\mu/(\mu+1)}.
\end{align}

%
%


Next, we consider the special case that $\alpha \rightarrow \infty$. From the definition of max divergence, we have
\begin{align}
	D_\infty (Pr_{\mathcal{M}}(D) \Vert Pr_{\mathcal{M}}(D'))
	=\sup_{Pr_{\mathcal{M}}(D)} \log \frac{p_\mathcal{M}(D)}{p_\mathcal{M}(D')} .
\end{align}

Refering to Equation \ref{equation:wdp2rdp_pre_ine_log}, we have
\begin{align}
	D_\infty (Pr_{\mathcal{M}}(D) \Vert Pr_{\mathcal{M}}(D'))
	\leq \sup_{Pr_{\mathcal{M}}(D)}  L \vert p_{\mathcal{M}}(D) - p_\mathcal{M}(D') \vert = L \Delta_p f .
\end{align}

Refering to Equation \ref{equation:wdp2rdp_most_important} , we know that
\begin{align}
	D_\infty (Pr_{\mathcal{M}}(D) \Vert Pr_{\mathcal{M}}(D'))
	\leq L \varepsilon^{\mu/(\mu+1)} .
\end{align}

Through the same methods, we can also prove that 
\begin{align}
	D_\infty (Pr_{\mathcal{M}}(D') \Vert Pr_{\mathcal{M}}(D)) \leq L  \varepsilon^{\mu/(\mu+1)}.
\end{align}

\subsection{Proof of Proposition 12}
\noindent\textbf{Proposition 12} (Post-Processing). Let $\mathcal{M}:\mathcal{D}\rightarrow \mathcal{R}$ be a $(\mu,\varepsilon)$-Wasserstein differentially private algorithm, and $\mathcal{G}: \mathcal{R} \rightarrow \mathcal{R}'$ be an arbitrary randomized mapping. For any order $\mu\in[1,\infty)$ and all measurable subsets $S\subseteq\mathcal{R}$, $\mathcal{G}(\mathcal{M})(\cdot)$ is also $(\mu,\varepsilon)$-Wasserstein differentially private, namely
\begin{align}
	W_\mu \left(Pr[\mathcal{G}(\mathcal{M}(D))\in S], Pr[\mathcal{G}(\mathcal{M}(D'))\in S]\right)\le \varepsilon.
\end{align}

\noindent\textit{proof.} 
Let $T=\{x \in \mathcal{R}: \mathcal{G}(x)\in S\}$, then we have

\begin{align}
	W_\mu (Pr[\mathcal{G}(\mathcal{M}(D))\in S], Pr[\mathcal{G}(\mathcal{M}(D'))\in S]
	&=W_\mu\left(
	Pr[\mathcal{M}(D)\in T], Pr[\mathcal{M}(D')\in T]
	\right)
	\\
	&= W_{\mu} (Pr_\mathcal{M}(D),Pr_\mathcal{M}(D')) \leq \varepsilon .
\end{align}

\subsection{Proof of Proposition 13}

\noindent\textbf{Proposition 13} (Group Privacy). Let $\mathcal{M} : \mathcal{D} \mapsto \mathcal{R}$ be a $(\mu, \varepsilon)$-Wasserstein differentially private algorithm. Then for any pairs of datasets $D, D' \in \mathcal{D}$ differing in $k$ data entries $x_1',\cdots,x_k'$ for any $i = 1, \cdots, k, \mathcal{M}(D)$ is
$(\mu, k \varepsilon)$-Wasserstein differentially private.

\noindent\textit{Proof.} We decompose the group privacy problem and denote $D,D_1'$ as a pair of adjacent datasets only differ in $x_1'$. Similarly, we denote $D_1'$ and $D_2'$, $D_2'$ and $D_3'$, $\cdots$, $D_{k-1}'$ and $D'$ as other $k-1$ pairs of adjacent datasets only differ in $x_2',x_3',\cdots, x_k'$ respectively. 

Recall that WDP satisfies triangle inequality in Proposition 2, then we have
\begin{equation}
	\begin{aligned}
		W_{\mu}(Pr_\mathcal{M}(D),Pr_\mathcal{M}(D')) \leq  W_\mu(Pr_\mathcal{M}(D), Pr_\mathcal{M}(D_1')) &+ W_\mu (Pr_\mathcal{M}(D_1'),Pr_\mathcal{M}(D_2'))+\cdots \\ &+ W_\mu (Pr_\mathcal{M}(D_{k-2}'),Pr_\mathcal{M}(D_{k-1}')) \\&+W_\mu (Pr_\mathcal{M}(D_{k-1}'),Pr_\mathcal{M}(D')) = k \varepsilon.
	\end{aligned}
\end{equation}

\subsection{Proof of Theorem 1}

\noindent\textbf{Theorem 1} (Advanced Composition) Suppose a randomized algorithm $\mathcal{M}$ consists of a sequence of $(\mu,\varepsilon)$-WDP algorithms $\mathcal{M}_1, \mathcal{M}_2\cdots,\mathcal{M}_T$, which perform on dataset $D$ adaptively and satisfy $\mathcal{M}_t: \mathcal{D}\rightarrow \mathcal{R}_t$, $t\in\{1,2,\cdots,T\}$. $\mathcal{M}$ is generalized $(\mu,\varepsilon)$-Wasserstein differentially private with $\varepsilon> 0$ and $\mu\geq1$ if for any two adjacent datasets $D,D^\prime\in\mathcal{D}$ hold that
\begin{equation}
	{ \exp
		\left[ 
		\beta 
		\sum_{t=1}^T 
		\mathbb{E}(
		W_\mu(Pr_{\mathcal{M}_t}(D),Pr_{\mathcal{M}_t}(D^\prime)))-
		\beta 
		\varepsilon
		\right]
	}    \leq \delta .  
\end{equation}
Where $\beta$ is a customization parameter that satisfies $\beta>0$.

\noindent\textit{Proof.} With definition of generalized $(\mu,\varepsilon)$-WDP, we have
\begin{align}
	Pr\left[ W_\mu(Pr_\mathcal{M}(D),Pr_\mathcal{M}(D^\prime))\geq \varepsilon  \right]
	&\leq Pr\left[ \beta \sum_{t=1}^T W_\mu(Pr_{\mathcal{M}_{t}}(D),Pr_{\mathcal{M}_{t}}(D^\prime)) \ge   \beta \varepsilon \right]  \label{equation:advanced_sequential_composition:triangle_inequality}
	\\&\le   \frac{\mathbb{E} \left[ \exp(  \beta    \sum_{t=1}^T W_\mu(Pr_{\mathcal{M}_t}(D),Pr_{\mathcal{M}_t}(D^\prime)))\right]}{\exp\left(\beta \varepsilon\right)} \label{equation:advanced_sequential_composition:markov_inequality}
	\\&\le \frac{ \exp  \left[ \beta \mathbb{E} \sum_{t=1}^T( W_\mu(Pr_{\mathcal{M}_t}(D),Pr_{\mathcal{M}_t}(D^\prime)))\right]}{\exp\left(
		\beta \varepsilon\right)} \label{equation:advanced_sequential_composition:jensen_inequality}
	\\&= \frac{ \exp\left[\beta \sum_{t=1}^T \mathbb{E}( W_\mu(Pr_{\mathcal{M}_t}D),Pr_{\mathcal{M}_t}(D^\prime)))\right]}
	{\exp\left(\beta \varepsilon\right)} \label{equation:advanced_sequential_composition:expectation}
	\\&= { \exp\left[ \beta \sum_{t=1}^T \mathbb{E}(  W_\mu(Pr_{\mathcal{M}_t}(D),Pr_{\mathcal{M}_t}(D^\prime)))- \beta \varepsilon\right]}    . \label{equation:advanced_sequential_composition:main}
\end{align}
Where Equation \ref{equation:advanced_sequential_composition:triangle_inequality} holds because triangle inequality (see Proposition 2) ensures that 
\begin{align}
	\sum_{t=1}^{T} W_\mu(Pr_{\mathcal{M}_{t}}(D),Pr_{\mathcal{M}_{t}}(D^\prime))\ge W_\mu(Pr_\mathcal{M}(D),Pr_\mathcal{M}(D^\prime)) . \nonumber
\end{align}
Inequality \ref{equation:advanced_sequential_composition:markov_inequality} holds because of Markov's inequality 
\begin{equation}
	Pr(\vert \cdot \vert \ge c)\le \frac{\mathbb{E}(\varphi(\vert \cdot \vert))}{\varphi(c)}, c>0. \label{equation:Markov_inequality}
\end{equation}
Here $\varphi(\cdot)$ can be any monotonically increasing function and satisfies the non-negative property. To simplify the computation of privacy budgets in WDP, we set $\varphi(\cdot)$ as  $\exp({ \cdot})$. 
Inequality \ref{equation:advanced_sequential_composition:jensen_inequality} holds because of Jensen's inequality. Equation \ref{equation:advanced_sequential_composition:expectation} is supported by the operational property of expectation. Thus, we find that Equation \ref{equation:advanced_sequential_composition:main} $\leq \delta$ implies $Pr\left[ W_\mu(\mathcal{M}(D),\mathcal{M}(D^\prime))\geq \varepsilon  \right]\leq \delta$.

\subsection{Proof of Theorem 2}

\noindent \textbf{Theorem 2} Suppose an algorithm $\mathcal{M}$ consists of a sequence of private algorithms $\mathcal{M}_1, \mathcal{M}_2\cdots,\mathcal{M}_T$ protected by Gaussian mechanism and satisfying $\mathcal{M}_t: \mathcal{D} \rightarrow \mathcal{R}$, $t=\left\{1,2,\cdots,T\right\}$. 
If the subsampling probability, scale parameter and $l_2$-sensitivity of algorithm $\mathcal{M}_t$ are represented by $q\in [0,1]$, $\sigma>0$ and $d_t\geq 0$, then the privacy loss under WDP at epoch $t$ is
\begin{equation}
	\begin{aligned}
		&W_{\mu} \left(Pr_{\mathcal{M}_t}(D), Pr_{\mathcal{M}_t}(D')\right)=
		\inf_{d_t}  \left[ \sum_{i=1}^n 
		\mathbb{E}\left(
		\vert Z_{ti} \vert^\mu
		\right) \right]^\frac{1}{\mu} ,
		\\ 
		&Z_t \sim\mathcal{N}\left(q d_t,(2-2q+2q^2)\sigma^2\right) .
	\end{aligned}
\end{equation}
Where $Pr_{\mathcal{M}_t}(D)$ is the outcome distribution when performing $\mathcal{M}$ on $D$ at epoch $t$. $d_t=\Vert g_t - g_t'\Vert_2$ represents the $l_2$ norm between pairs of adjacent  gradients $g_t$ and $g_t'$.
In addition, $Z_t$ is a vector follows Gaussian distribution, and $Z_{ti}$ represents the $i$-th component of $Z_t$.

\noindent\textit{Proof.} With Gaussian mechanism in a subsampling scenario, we have
\begin{align}
	&Pr_{\mathcal{M}_t}(D) =	(1-q) \mathcal{N}(0,\sigma^2) + q\mathcal{N}(d_t, \sigma^2) \nonumber,\\
	&Pr_{\mathcal{M}_t}(D') = \mathcal{N}(0,\sigma^2). \nonumber
\end{align}

To facilitate the later proof, we slightly simplify the expression of $\mathcal{M}_t(D)$.
\begin{align}
	Pr_{\mathcal{M}_t}(D) &= (1-q) \mathcal{N}(0,\sigma^2) + q\mathcal{N}(d_t, \sigma^2)
	\\
	&=\mathcal{N}\left(0,(1-q)^2\sigma^2\right) + \mathcal{N}\left(q d_t, q^2 \sigma^2\right)
	\\ 
	&=\mathcal{N}\left(q d_t,(1-2q+2q^2)\sigma^2\right).
\end{align}

Then we compute the privacy loss at epoch $t$
\begin{align}
	W_\mu \left(Pr_{\mathcal{M}_t}(D), Pr_{\mathcal{M}_t}(D')\right) 
	= 
	\inf_{X\sim Pr_{\mathcal{M}_t}(D)\atop Y\sim Pr_{\mathcal{M}_t}(D')}  \left[ \mathbb{E}\; \Vert 
	X_t - Y_t\Vert^\mu \right]^\frac{1}{\mu} .
\end{align}

Let $Z_t = X_t - Y_t$, thus we have
\begin{align}
	Z_t \sim \mathcal{N}\left(q d_t, 2 -2 q + 2q^2\right) .
\end{align}

The privacy loss is 
\begin{align}
	W_\mu  \left(Pr_{\mathcal{M}_t}(D), Pr_{\mathcal{M}_t}(D')\right)
	=
	\inf_{d_t}  \left[ \mathbb{E}  \left(    \Vert
	Z_t
	\Vert^\mu \right)  \right]^\frac{1}{\mu} .
\end{align}

Refering to the definition of norm, we can obtain
\begin{align}
	\Vert Z_t \Vert= \left( \sum_{i=1}^n \vert Z_{ti} \vert^\mu\right)^{\frac{1}{\mu}}
	\Rightarrow
	\;
	\Vert Z_t \Vert^\mu
	=
	\sum_{i=1}^n \vert Z_{ti} \vert^\mu .
\end{align}

According to the summation property of expectation, we have
\begin{align}
	\mathbb{E}\left[ 
	\Vert Z_t \Vert^\mu
	\right]
	=
	\mathbb{E}\left[
	\sum_{i=1}^n \vert Z_{ti} \vert^\mu
	\right]
	=
	\sum_{i=1}^n 
	\mathbb{E}\left(
	\vert Z_{ti} \vert^\mu
	\right) .
\end{align}

Finally, we have
\begin{align}
	W_\mu  \left(Pr_{\mathcal{M}_t}(D), Pr_{\mathcal{M}_t}(D')\right)
	=
	\inf_{d_t}  \left[ 
	\sum_{i=1}^n 
	\mathbb{E}\left(
	\vert Z_{ti} \vert^\mu
	\right)
	\right]^\frac{1}{\mu} .
\end{align}

\subsection{Proof of Theorem 3}

\noindent \textbf{Theorem 3} (Tail bound) Under the conditions described in Theorem 2, $\mathcal{M}$ satisfies $(\mu,\delta)$-WDP for 
\begin{equation}
	\begin{aligned}
		&	\log \delta = \beta \sum_{t=1}^T \inf_{d_t}	\left[  \sum_{i=1}^{n}
		\mathbb{E}
		\left(
		\vert
		Z_{ti}
		\vert^\mu  
		\right) 
		\right]^\frac{1}{\mu}   - \beta \varepsilon, \\
		&Z \sim	\mathcal{N}\left(q d_t,(2-2q+2q^2)\sigma^2\right) .
	\end{aligned}
\end{equation}

\noindent\textit{Proof.} In Theorem 1, we have proved that
\begin{equation}\label{cite_theorem_1}
	{ \exp
		\left[ 
		\beta 
		\sum_{t=1}^T 
		\mathbb{E}(
		W_\mu(Pr_{\mathcal{M}_t}(D),Pr_{\mathcal{M}_t}(D^\prime)))-
		\beta 
		\varepsilon
		\right]
	}    \leq \delta .  
\end{equation}

Taking logarithms on both sides of Equation \ref{cite_theorem_1}, we can obtain
\begin{align}\label{equation:log_delta}
	\beta  \sum_{t=1}^T\mathbb{E}( W_\mu(
	Pr_{\mathcal{M}_t}(D), Pr_{\mathcal{M}_t}(D^\prime)
	)) - \beta \varepsilon \leq \log \delta .
\end{align}

In Theorem 2, we have proved that 
\begin{equation}\label{cite_theorem_2}
	\begin{aligned}
		W_{\mu} \left(Pr_{\mathcal{M}_t}(D),Pr_{\mathcal{M}_t}(D')\right)=
		\inf_{d_t}  \left[ \sum_{t=1}^T 
		\mathbb{E}\left(
		\vert Z_{ti} 
		\vert^\mu
		\right) \right]^\frac{1}{\mu} ,
	\end{aligned}
\end{equation}
where $Z \sim	\mathcal{N}\left(q d_t,(2-2q+2q^2)\sigma^2\right)$ and $d_t = \Vert g_t - g_t'\Vert_2$. 

Plugging Equation \ref{cite_theorem_2} into Equation \ref{equation:log_delta}, we can obtain
\begin{align}\label{equation:proof_of_theorem3_prototype}
	\beta \sum_{t=1}^T\mathbb{E} \left[ 
	\inf_{d_t}  \left[ \sum_{i=1}^n 
	\mathbb{E}\left(
	\vert Z_{ti} \vert^\mu
	\right) \right]^\frac{1}{\mu}  \right] - \beta \varepsilon\leq log \delta.
\end{align}
Where $\mathbb{E} \left(\vert
Z
\vert^\mu\right)$ can be obtained with the help of Lemma 1, thus we regard it as a computable whole part. 

Observing Equation \ref{equation:proof_of_theorem3_prototype}, we find that the uncertainty comes from two parts: Gaussian random variable $Z$ and the norm of pairwise gradients $\Vert g_t - g_t'\Vert_2$. However, these two uncertainties have been eliminated by the inner expectation and the operation of infimum. Thus, we no longer need outside $\mathbb{E}$ and the expression can be simplified as 
\begin{align}
	\beta \sum_{t=1}^T \inf_{d_t}	\left[  
	\sum_{i=1}^{n}
	\mathbb{E}
	\left(
	\vert
	Z_{ti}
	\vert^\mu  
	\right) 
	\right]^\frac{1}{\mu}  - \beta \varepsilon\leq log \delta.
\end{align}


We always want the probability of failure to be as small as possible, thus we replace the unequal sign with the equal sign as follow
\begin{align}
	\log \delta = \beta \sum_{t=1}^T \inf_{d_t}	\left[ 
	\sum_{i=1}^{n}
	\mathbb{E}
	\left(
	\vert
	Z_{ti}
	\vert^\mu  
	\right) 
	\right]^\frac{1}{\mu}   - \beta \varepsilon .
\end{align}


\subsection{Proof of Lemma 1}

\noindent \textbf{Lemma 1} (Raw Absolute Moment) Assume that $Z \sim\mathcal{N}(q d_t,(2-2q+2q^2) \sigma^2)$, we can obtain the raw absolute moment of $Z$ as follow
\begin{align}
	\mathbb{E} \left(\vert Z \vert^\mu \right) = 
	\left( 2 Var\right)^{\frac{\mu}{2}}\frac{
		GF\left({\frac{\mu+1}{2}}\right)
	}{\sqrt{\pi}} \mathcal{K} \left(
	-\frac{\mu}{2}, \frac{1}{2}; -\frac{ q^2 d_t^2}{2 Var}
	\right) . \nonumber
\end{align}
Where $Var$ represents the Variance of Gaussian random variable $Z$, and can be expressed as $Var=(2-2q+2q^2) \sigma^2$.
$GF\left({\frac{\mu+1}{2}}\right)$ represents Gamma function as  
\begin{align}
	GF\left({\frac{\mu+1}{2}}\right) = \int_0^\infty x^{{\frac{\mu+1}{2}}-1} e^{-x} dx,
\end{align}
and $\mathcal{K} \left(
-\frac{\mu}{2}, \frac{1}{2}; -\frac{ q^2 d_t^2}{2 Var}
\right)$ represents Kummer's confluent hypergeometric function as
\begin{align}
	\sum_{n=0}^\infty 
	\frac{
		{
			q^{2n} d_t
		}^{2n}
	}
	{
		n! \cdot 
		4^n (1-q+q^2)^n \sigma^{2n}
	}
	\prod_{i=1}^n  \frac{
		\mu-2i+2
	}{
		1+2i-2
	}. 
\end{align}

\noindent\textit{Proof.} From Equation 17 in \citet{winkelbauer2012moments}, we can obtain the expression of $\mathbb{E} \left(\vert Z \vert^\mu \right)$ as follow
\begin{align}
	\mathbb{E} \left(\vert Z \vert^\mu \right) =\left( 2 Var\right)^{\frac{\mu}{2}} \frac{
		GF\left({\frac{\mu+1}{2}}\right)
	}{\sqrt{\pi}} \mathcal{K} \left(
	-\frac{\mu}{2}, \frac{1}{2}; -\frac{q^2 d_t^2}{2 Var}
	\right) . 
\end{align}
Where $\mathcal{K} \left(
-\frac{\mu}{2}, \frac{1}{2}; -\frac{ q^2 d_t^2}{2 Var}
\right) $ deduce further as follows
\begin{align}
	\mathcal{K} \left(
	-\frac{\mu}{2}, \frac{1}{2}; -\frac{ q^2 d_t^2}{2 Var}
	\right)  
	&= 
	\mathcal{K} \left(
	-\frac{\mu}{2}, \frac{1}{2}; -\frac{ q^2 d_t^2}{2  (2-2q+2q^2)\sigma^2}
	\right)  
	\\
	&= 
	\mathcal{K} \left(
	-\frac{\mu}{2}, \frac{1}{2}; -\frac{ q^2 d_t^2}{4  (1-q+q^2)\sigma^2}
	\right)  
	\\
	&= \sum_{n=0}^\infty \frac{
		\left(-\frac{\mu}{2}\right)^{\overline{n}} 
	}{
		\left(\frac{1}{2}\right)^{\overline{n}}
	} 
	\frac{\left( 
		-\frac{ q^2 d_t^2}{4  (1-q+q^2)\sigma^2}
		\right)^n}{n!} 
	\\
	&= \sum_{n=0}^\infty 
	\frac{
		\left(-\frac{\mu}{2}\right)^{\overline{n}} 
	}{
		\left(\frac{1}{2}\right)^{\overline{n}}
	} 
	\left(-1\right)^n
	\frac{\left( 
		\frac{ q^2 d_t^2}{4  (1-q+q^2)\sigma^2}
		\right)^n}{n!} 
	\\
	&= \sum_{n=0}^\infty 
	\left(-1\right)^n
	\frac{
		\left(-\frac{\mu}{2}\right)^{\overline{n}} 
	}{
		\left(\frac{1}{2}\right)^{\overline{n}}
	} 
	\frac{\left( 
		q^2 d_t^2
		\right)^{n}}{
		n! \cdot 
		\left({4  (1-q+q^2)\sigma^2}\right)^{n}
	}\\
	&= \sum_{n=0}^\infty 
	\left(-1\right)^n
	\frac{
		\left(-\frac{\mu}{2}\right)^{\overline{n}} 
	}{
		\left(\frac{1}{2}\right)^{\overline{n}}
	} 
	\frac{
		{
			q^{2n} d_t
		}^{2n}
	}
	{
		n! \cdot 
		4^n (1-q+q^2)^n \sigma^{2n}
	}. 
	\label{equation:kummer_last}
\end{align}
Where $(-\frac{\mu}{2})^{\overline{n}}$ is the rising factorial of $-\frac{\mu}{2}$ (see \citet{winkelbauer2012moments})
\begin{align}
	\left( -\frac{\mu}{2} \right)^{\overline{n}} &=	\frac{GF(-\frac{\mu}{2}+n)}{GF(-\frac{\mu}{2})}\\
	&=\left(-\frac{\mu}{2}\right)\cdot 
	\left(-\frac{\mu}{2} + 1\right)  \cdot
	...  \cdot
	\left(-\frac{\mu}{2} + n-1\right)\\
	&=\left(-1\right)^n 
	\left(\frac{1}{2}\right)^n \mu \cdot  \left(\mu-2\right)\cdot ...\cdot \left(   
	\mu -2n + 2
	\right) . 
	\label{equation:rising_factorial_1_last}
\end{align}
\begin{align}
	\left(\frac{1}{2} \right)^{\overline{n}} &= \frac{GF(\frac{1}{2}+n)}{GF(\frac{1}{2})}\\
	&=\left(\frac{1}{2}\right)\cdot 
	\left(\frac{1}{2} + 1\right)  \cdot
	...  \cdot
	\left(\frac{1}{2} + n-1\right)
	\\
	&=\left(\frac{1}{2}\right)^n \left[1\cdot 3\cdot ... \cdot (1+2n-2)\right] . 
	\label{equation:rising_factorial_2_last}
\end{align}
From Equation \ref{equation:rising_factorial_1_last} and \ref{equation:rising_factorial_2_last}, we have
\begin{align}
	\frac{
		\left( -\frac{\mu}{2} \right)^{\overline{n}} 
	}{
		\left( \frac{1}{2} \right)^{\overline{n}}
	}
	&=\frac{ (-1)^n \cdot
		\mu \cdot  \left(\mu-2\right)\cdot ...\cdot \left(   
		\mu -2n + 2
		\right)
	}{
		\left[1\cdot 3\cdot ... \cdot (1+2n-2)\right]
	}
	\\
	&=(-1)^n \frac{
		\prod_{i=1}^n \mu-2(i-1)
	}{
		\prod_{i=1}^n 1+2i-2
	} 
	\\&
	=(-1)^n\prod_{i=1}^n  \frac{
		\mu-2i+2
	}{
		1+2i-2
	} .
	\label{equation:rising_factorial_result}
\end{align}
Combing Equation \ref{equation:kummer_last} and \ref{equation:rising_factorial_result}, we can obtain
\begin{align}
	\mathcal{K} \left(
	-\frac{\mu}{2}, \frac{1}{2}; -\frac{ q^2 d_t^2}{2 Var}
	\right) 
	&= 
	\sum_{n=0}^\infty 
	\frac{
		{
			q^{2n} d_t
		}^{2n}
	}
	{
		n! \cdot 
		4^n (1-q+q^2)^n \sigma^{2n}
	}
	\prod_{i=1}^n  \frac{
		\mu-2i+2
	}{
		1+2i-2
	} 
	\\
	&= \sum_{n=0}^\infty 
	\frac{
		{
			q^{2n} d_t
		}^{2n}
	}
	{
		n! \cdot 
		4^n (Var)^n
	}
	\prod_{i=1}^n  \frac{
		\mu-2i+2
	}{
		1+2i-2
	} .
\end{align}
%
%

\section{Experiments}

\subsection{Composition with Clipping}

\begin{figure}[htbp]
	\centering	
	\subfigure[0.05-quantile of $\Vert g_t \Vert$]{
		\begin{minipage}[t]{0.24\linewidth}
			\centering
			\includegraphics[width=1.4in]{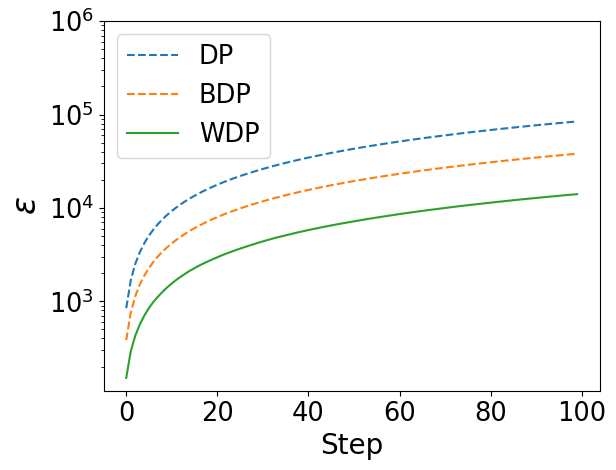}
		\end{minipage}
	}%
	\subfigure[0.50-quantile of $\Vert g_t \Vert$]{
		\begin{minipage}[t]{0.24\linewidth}
			\centering
			\includegraphics[width=1.4in]{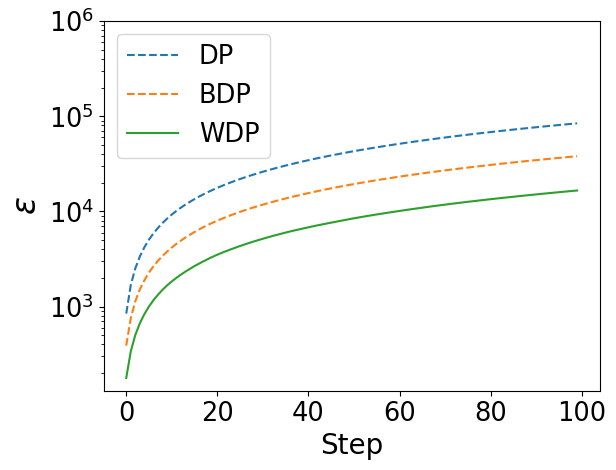}
		\end{minipage}
	}%
	\subfigure[0.75-quantile of $\Vert g_t \Vert$]{
		\begin{minipage}[t]{0.24\linewidth}
			\centering
			\includegraphics[width=1.4in]{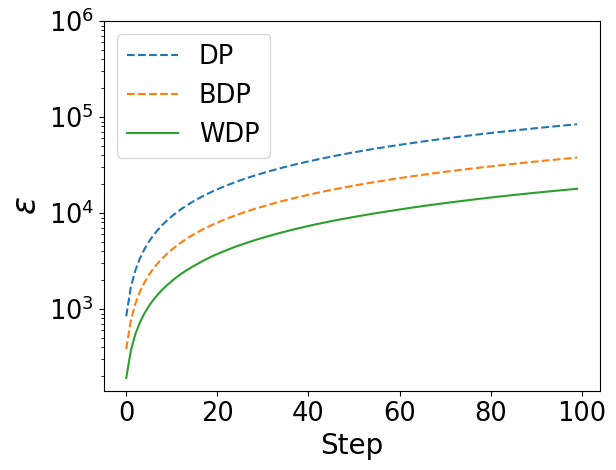}
		\end{minipage}
	}%
	\subfigure[0.99-quantile of $\Vert g_t \Vert$]{
		\begin{minipage}[t]{0.24\linewidth}
			\centering
			\includegraphics[width=1.4in]{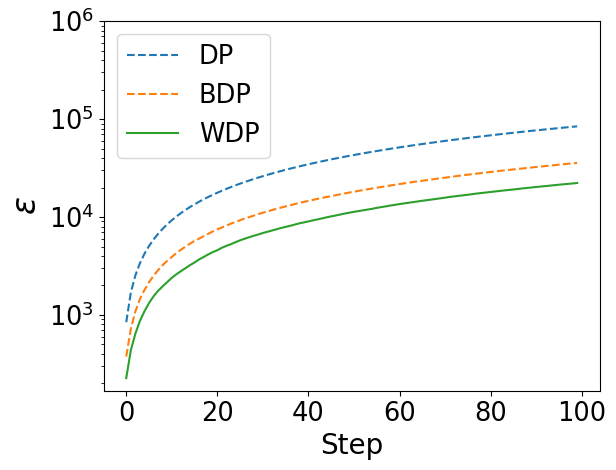}
		\end{minipage}
	}%
	\centering
	\caption{Privacy budgets over synthetic gradients obtained by moments accountant under DP, Bayesian accountant under BDP and Wasserstein accountant under WDP when applying gradient clipping.}
	\label{fig:composition_clip}
\end{figure}

Figure \ref{fig:composition_clip} demonstrate the changing process of privacy budget as step increases. 
We find that the impact of $C$ on the privacy budget has decreased, because the gradient norm is limited by the clipping threshold, and the gap of privacy budgets between different DP frameworks has narrowed. 
However, this does not affect WDP to still get the lowest cumulative privacy budget, and this value grows the a bit slower than that of DP and BDP.

%
%

\section{Several Basic Concepts in Differential Privacy}


\textbf{Definition 6} (Differential Privacy~\cite{DBLP:conf/tcc/DworkMNS06}). A randomized algorithm $\mathcal{M}:\mathcal{D}\rightarrow\mathcal{R}$ is $\left(\varepsilon,\delta\right)$-differentially private if for any adjacent datasets $D,D’\in \mathcal{D}$ and all measurable subsets $S\subseteq\mathcal{R}$ the following inequality holds: 
\begin{equation}
	Pr\left[\mathcal{M}\left(D\right)\in S\right]\le e^\varepsilon Pr\left[\mathcal{M}\left(D'\right)\in S\right]+\delta .  \label{equation:differential_privacy}
\end{equation}
Where $Pr[\cdot]$ is the notation of probability, and $\varepsilon$ is known as the privacy budget. 
In particular, if $\delta=0$, $\mathcal{M}$ is said to preserve $\varepsilon$-DP or pure DP.

\noindent\textbf{Definition 7} (Privacy Loss of  DP). For a randomized algorithm $\mathcal{M}: \mathcal{D}\rightarrow \mathcal{R}$, and $o$ is the outcome of algorithm $\mathcal{M}$, then the privacy loss of the $\mathcal{M}$ can be defined as
\begin{equation}
	Loss\left(o\right)=\log\frac{Pr\left[\mathcal{M}\left(D\right)=o\right]}{Pr\left[\mathcal{M}\left(D^\prime\right)=o\right]}.
\end{equation}
Privacy budget is the strict upper bound of privacy loss in $\varepsilon$-differential privacy, and is a quasi upper bound of privacy loss with the confidence of $1-\delta$ in ($\varepsilon$,$\delta$)-differential privacy.

\noindent\textbf{Definition 8} ($l_p$-Sensitivity \cite{DBLP:conf/stoc/DworkL09}).  Sensitivity in DP theory can be defined by maximum $p$-norm distance between the same query functions of two adjacent datasets ${D}$ and ${D}'$
\begin{equation}
	{\Delta}_p f=\sup_{\rho({D},{D'})\le 1} \Vert f({D})-f({D'})\Vert_p .
\end{equation}
Where $f:\mathcal{D}\rightarrow \mathbb{R}^d$ is a $d$-dimension query function, $\rho({D},{D'})=\Vert D-D' \Vert_p$ is the norm function between ${D}$ and ${D'}$. $l_p$-sensitivity measures the largest difference between all possible adjacent datasets. 

\noindent\textbf{Definition 9} (Rényi Differential Privacy \cite{DBLP:conf/csfw/Mironov17}). A randomized algorithm $\mathcal{M}:\mathcal{D}\rightarrow\mathcal{R}$ is said to preserve $\left(\alpha,\varepsilon\right)$-RDP if for any adjacent datasets $D,D^\prime \in \mathcal{D}$ the following holds
\begin{equation}
	\begin{aligned}
		D_\alpha (Pr_\mathcal{M}(D)\Vert Pr_\mathcal{M}(D'))=\frac{1}{\alpha-1} \log\mathbb{E}_{o\sim\mathcal{M}\left(D^\prime\right)}
		\left[\left(\frac{p_{\mathcal{M}\left({D}\right)}\left(o\right)}{p_{{\mathcal{M}}\left({D}^\prime\right)}\left(o\right)}\right)^\alpha\right]\le\varepsilon .
	\end{aligned}\label{equation:renyi_privacy}
\end{equation}
Where $\alpha\in\left(1,+\infty\right)$ is the order of RDP, $o$ is the output of algorithm $\mathcal{M}$. $Pr_\mathcal{M}(D)$ and $Pr_\mathcal{M}(D')$ are probability distributions, while $p_\mathcal{M}(D)$ and $p_\mathcal{M}(D')$ are probability density functions.

\noindent\textbf{Definition 10} (Strong Bayesian Differential Privacy \cite{DBLP:conf/icml/TriastcynF20}). A randomized algorithm $\mathcal{M}:\mathcal{D}\rightarrow\mathcal{R}$ is said to satisfy $\left(\varepsilon_b,\delta_b\right)$-strong Bayesian differential privacy if for any adjacent datasets $D,D’\in\mathcal{D}$ the following holds
\begin{equation}
	Pr\left[\log \frac{p(o\vert D)}{p(o\vert D^\prime)}\ge \varepsilon_b\right] \le \delta_b.
\end{equation}
Where $\varepsilon_b$ and $\delta_b$ are privacy budget and failure probability in BDP  \cite{DBLP:conf/icml/TriastcynF20}. 
Where $o$ is the output satisfying $o=\mathcal{M}(\cdot)$. $p(o|D)$ and $p(o|D')$ are probability density functions of adjacent datasets.

\noindent\textbf{Definition 11} (Bayesian Differential Privacy \cite{DBLP:conf/icml/TriastcynF20}).
Suppose the only different data entry $x'$ follows a certain distribution $b(x)$, namely $x'\sim b(x)$. A randomized algorithm $\mathcal{M}:\mathcal{D}\rightarrow\mathcal{R}$ is said to satisfy $\left(\varepsilon_b,\delta_b\right)$-Bayesian differential privacy if for any neighboring datasets $D,D'\in \mathcal{D}$ and any set of outcomes $\mathcal{O}$ the following holds
\begin{equation}
	Pr\left[\mathcal{M}\left(D\right)\in \mathcal{O} \right]\le e^{\varepsilon_b}Pr\left[\mathcal{M}\left(D\right)\in \mathcal{O} \right]+\delta_b      \label{equation:Bayesian_differential_privacy}.
\end{equation}
From the above definitions, we find that strong BDP is inspired by RDP, and the definition of BDP is similar to that of DP. 
Therefore, the weaknesses of DP, BDP and RDP are similar: 
(1) Their privacy losses do not satisfy symmetry and triangle inequality, which prevent them from becoming metrics. (2) Their privacy budgets tend to be overstated. 
To alleviate these problems, we propose Wasserstein differential privacy in this paper, expecting to achieve better properties in privacy computing, and thus obtain higher performances in private machine learning.

\end{document}